\documentclass{article}

\usepackage[preprint]{neurips_2026}

\usepackage[T1]{fontenc}
\usepackage[utf8]{inputenc}
\usepackage{microtype}

\usepackage{url}
\usepackage{hyperref}

\usepackage{amsmath}
\usepackage{amssymb}
\usepackage{amsfonts}
\usepackage{bm}
\usepackage{mathtools}

\usepackage{booktabs}
\usepackage{multirow}
\usepackage{array}

\usepackage{graphicx}
\usepackage{subcaption}

\usepackage[table]{xcolor}
\usepackage{nicefrac}
\usepackage{enumitem}

\usepackage{float}

\usepackage[normalem]{ulem}

\title{Allocation Before Ranking: Decoupled Token Compression for OmniLLMs}

\author{%
  \textbf{Zhenghui Guo}$^{1}$ \quad
  \textbf{Yilin Yang}$^{1}$ \quad
  \textbf{Yuanbin Man}$^{2}$ \quad
  \textbf{Miao Yin}$^{2}$ \\
  \textbf{Weidong Shi}$^{1}$ \quad
  \textbf{Rabimba Karanjai}$^{1}$ \quad
  \textbf{Omprakash Gnawali}$^{1}$ \quad
  \textbf{Chengming Zhang}$^{1}$ \\
  $^{1}$Department of Computer Science, University of Houston \\
  $^{2}$Department of Computer Science and Engineering,
  The University of Texas at Arlington
}

\begin{document}

\maketitle

\begin{abstract}
Token compression in OmniLLMs is typically posed as a single saliency-ranking problem: score each multimodal token, keep the top-$K$. We argue this abstraction is mis-specified. The same attention score simultaneously decides two things: how much retained capacity each modality receives, and which tokens
within a modality are kept. A shared top-$K$ rule therefore inherits this audio-favoring allocation prior, spending retained capacity on audio before video tokens have a chance to compete. We propose \textsc{Macer}, a training-free compressor that first assigns explicit audio and video budgets,
then performs allocation-normalized ranking within each modality at modality-specific shallow layers. \textsc{Macer} significantly reduces token cost while preserving accuracy across audio-grounded, audio--video joint, visual-dominant, and video-centric benchmarks. At 25\% retention, \textsc{Macer} preserves 98.7\% of full-token performance on Qwen2.5-Omni-7B and 97.3\% on Qwen2.5-Omni-3B. On Qwen2.5-Omni-7B, this 25\% setting reaches OmniZip-level performance at 45\% retention while using lower FLOPs. On OmniVinci-9B, the same allocation-before-ranking principle improves over shared top-$K$ ranking by up to 12.9 points.
\end{abstract}

\section{Introduction}
Omnimodal large language models (OmniLLMs) unify audio and video understanding by
interleaving both modalities and processing them through one decoder~\citep{xu2025qwen25omni,
ye2025omnivinci,fu2024vita,ge2025arc}. This design enables a single model to jointly understand audio and video, but it also makes inference scale with the combined multimodal sequence. Token compression is therefore necessary. Most existing compressors treat this problem as one saliency ranking: assigning a score to each multimodal token and keeping the top-scoring tokens under a fixed budget. We argue that this ranking-only abstraction is incomplete for OmniLLM token compression.

The failure is ``\emph{one score, two decisions}.'' In a shared decoder, attention magnitude both allocates probability mass across modalities and orders tokens within each modality. A shared top-$K$ rule therefore uses the same score to decide \emph{how much} audio/video capacity survives and \emph{which} audio/video tokens survive. These are different decisions: cross-modal capacity allocation should be controlled before within-modality token selection, not induced accidentally by raw shared-attention mass. In OmniLLM, the coupling is concrete: shared attention exhibits a persistent, content-responsive audio allocation prior, so ranking audio and video tokens in one space can spend retained capacity on audio before visual evidence has a fair chance to compete.

Our diagnostics give three observations that motivate the method. \textbf{Observation 1: shared attention is not a neutral cross-modal saliency space.} Even under video-essential queries, audio receives higher attention mass per token across layers; null-content and distance controls show that the effect is content-responsive rather than a fixed positional sink. \textbf{Observation 2: the audio-favoring bias is a budgeting problem, not a video-selection problem.} ToMe-style~\citep{bolya2023tome} video merging and random video dropping produce similar audio-mass trajectories at matched video keep ratio, while changing the retained audio share directly changes accuracy. \textbf{Observation 3: audio and video become selection-readable at different shallow depths.} Audio readout is stable early, whereas video readout is more layer-sensitive and improves around later shallow layers.

We propose \textsc{Macer}, a training-free inference-time compressor that instantiates this allocation-selection principle. \textsc{Macer} first assigns explicit audio and video budgets, preventing retained capacity from being implicitly determined by raw shared-attention mass. It then ranks tokens only within each modality using allocation-normalized attention scores. Because audio and video saliency become readable at different shallow depths, \textsc{Macer} reads them at modality-specific layers before pruning. A lightweight video-local temporal coverage term is applied only inside the assigned video budget to reduce temporal over-concentration.

We evaluated \textsc{Macer} on three shared-decoder OmniLLM backbones (Qwen2.5-Omni-7B, Qwen2.5-Omni-3B, and OmniVinci-9B) across four benchmarks that span audio-grounded, audio-video joint, visual-dominant, and video-centric tasks. \textsc{Macer} achieves the strongest aggregate performance among compared token-pruning baselines at every evaluated retention ratio. At 25\% retention, \textsc{Macer} preserves 98.7\% of full-token performance on Qwen2.5-Omni-7B (vs.\ 91.6\% for OmniZip). On Qwen2.5-Omni-3B, it preserves 97.3\% (vs.\ 95.5\% for OmniZip). On OmniVinci-9B, \textsc{Macer} outperforms shared top-$K$ ranking by up to 12.9 points, showing that the allocation--selection separation transfers across models.

\begin{figure*}[!t]
    \centering
    \includegraphics[width=\textwidth]{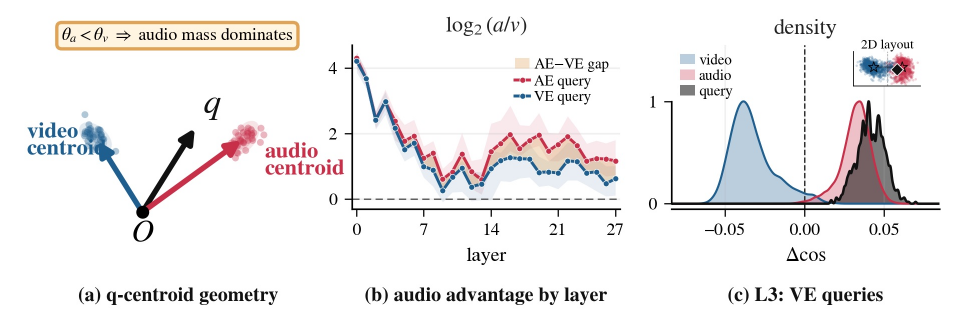}
    \vspace{-1.5em}
    \caption{\textbf{Shared attention favors audio across depth.}
    $c_a$ and $c_v$ denote the mean audio and video keys.
    \textbf{(a)} The query $q$ is closer to $c_a$ than to $c_v$, so shared attention favors audio.
    \textbf{(b)} Under both video-essential (VE) and audio-essential (AE) queries, $\log(\mathrm{mass}_a/\mathrm{mass}_v)$ stays positive across layers.
    \textbf{(c)} At L3 under VE queries, pooled query positions satisfy $\cos(q,c_a)>\cos(q,c_v)$.
    This geometry illustrates why raw shared attention is not a neutral token-saliency space for omnimodal compression.}
    \label{fig:teaser_audio_advantage}
    \vspace{-1em}
\end{figure*}

\begin{itemize}[leftmargin=1.2em, topsep=0.3em, itemsep=0.15em]

\item We identify a structural failure of global Top-$K$ omnimodal compression:
a single importance score determines both the audio-video token budget and the
tokens retained within each modality. No saliency signal fixes this coupling unless
the token budget is allocated explicitly before selection.

\item We propose \textsc{Macer}, a 
training-free compressor that decides how many audio and video tokens to keep before deciding which tokens to keep. \textsc{Macer} first allocates audio/video keep-ratios, then ranks tokens within each modality using attention normalized over that modality, and reads audio and video saliency at different shallow layers. A small temporal-coverage rule prevents retained video tokens from clustering in time. The method requires no training, no model changes, and no second forward pass.

\item Our results validate that \textsc{Macer} preserves 98.7\% / 97.3\% of full-token accuracy on 
Qwen2.5-Omni-7B/3B at 25\% retention, and improves over shared top-$K$ ranking by up to 12.9 points on 
OmniVinci-9B.
\end{itemize}
\section{Related work}
\label{sec:related}
\paragraph{Modality competition in multimodal LLMs.}
Prior work has shown that MLLMs do not use modalities uniformly: modality bias appears across vision, audio, time-series, and graph inputs~\citep{wu2025when,zheng2025mllms}, and audio-language or audio-visual analyses show that cross-modal attention can underuse task-relevant modalities~\citep{wang2025pay}. These works mainly diagnose representation or attention behavior. We study the deployment-time consequence for compression. 
When audio and video tokens share a decoder softmax under a fixed retained-token budget, imbalance becomes a capacity-allocation problem: the compressor must decide both which tokens are salient and how much capacity each modality receives.

\paragraph{Token compression for multimodal LLMs.}
Multimodal token-compression methods differ by where the retention decision is made and what signal drives it. 
Encoder-side methods reduce visual, audio-visual, or omnimodal context before or near decoder integration using encoder features, cross-modal guidance, or temporal/event structure~\citep{tao2025omnizip,ding2026omnisift,li2026dash,guo2026event,shao2025tokensurvey}. 
LLM-side methods instead prune or merge tokens using attention, redundancy, or sparsification signals inside the model~\citep{bolya2023tome,chen2024fastv,shang2025llavaprumerge,
yang2025visionzip,zhang2025sparsevlm,ye2025fitprune,xing2024pyramiddrop,huang2025prunevid,
tao2025dycoke,jiang2025storm,shao2025holitom}, though recent VLM studies question whether text-to-visual attention alone is reliable for visual-token importance~\citep{zhang2025vispruner}. 
These methods provide useful saliency signals, but the audio-video budget is usually fixed, implicit, or induced by a shared score. 
A single shared ranking therefore leaves capacity allocation coupled to within-modality token selection.

\paragraph{Omnimodal token compression.}
The closest omnimodal line compresses audio-video contexts directly. 
OmniZip~\citep{tao2025omnizip} uses an encoder-side audio-guided scheme before decoder integration, while OmniSIFT, DASH, and FastAV explore modality-asymmetric pruning, audio-driven chunking, or audio-visual pruning~\citep{ding2026omnisift,li2026dash,jung2026fastav}. 
MACER differs along the capacity-allocation axis: it exposes the retained audio share as an explicit deployment-level control variable, then performs allocation-normalized selection within each modality using shallow decoder-side attention at modality-specific layers. MACER is complementary to saliency-signal design: it targets a different axis, making the retained audio-video capacity split explicit before modality-local token selection.

\begin{figure*}[t]
    \centering
    \includegraphics[width=\textwidth]{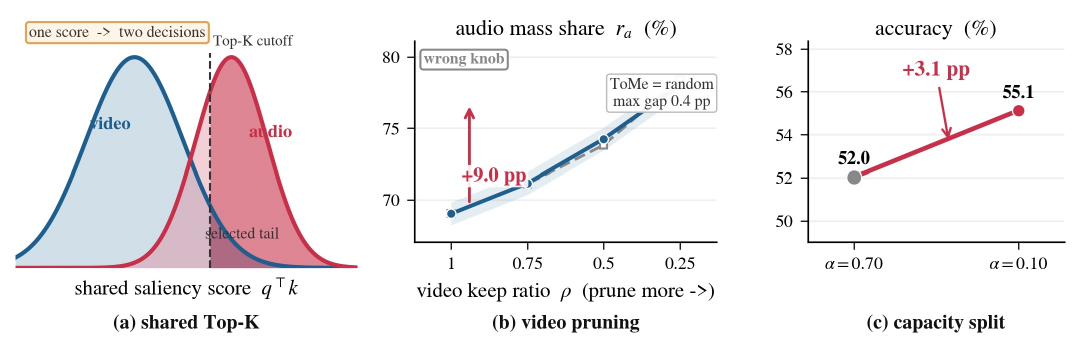}
    \vspace{-1.5em}
    \caption{\textbf{Capacity allocation, not within-video selection, controls cross-modal balance.}
    (a) A shared top-$K$ rule uses one score for two decisions.
    (b) ToMe and random dropping differ by at most 0.4 pp at matched video keep ratio,
    whereas reducing the video keep ratio itself shifts audio mass share by 9.0 pp.
    (c) The retained audio share $\alpha$ affects accuracy, with a video-heavy 
    split ($\alpha=0.10$) outperforming an audio-heavy split ($\alpha=0.70$) by
    3.1 points.}
    \vspace{-1em}
    \label{fig:allocation_vs_intra}
\end{figure*}

\section{Shared Ranking Couples Allocation and Selection}
\label{sec:mechanism}

Shared-budget omnimodal compression is often treated as a token-ranking problem: assign a saliency score to each audio and video token, then retain the top-scoring tokens under a common budget. This abstraction is incomplete for shared-decoder OmniLLMs. In a shared attention~\citep{vaswani2017attention}, attention magnitude does not only score token-level relevance; it also determines how probability mass is allocated across modalities. A shared ranking rule therefore contains a cross-modal allocation term even when the desired decision is within-modality token selection.

\paragraph{One attention score, two decisions.}
\label{sec:setup_decomp}
A query's attention to a token does two distinct things at once: it scores how relevant that token is \emph{within its modality}, and it decides how much probability mass goes to \emph{that modality} versus the other (Fig.~\ref{fig:allocation_vs_intra}a). An exact decomposition makes this concrete. Let each key decompose as $k_i^{(m)} = c_m + \epsilon_i^{(m)}$, where $c_m = \frac{1}{|m|}\sum_{i\in m} k_i^{(m)}$ is the per-clip modality centroid and $\epsilon_i^{(m)}$ is the within-modality residual. The log attention mass on modality $m$ admits the exact decomposition
\begin{equation}
\log \mathrm{mass}_m(q)
= \underbrace{q^{\!\top} c_m}_{\Delta_{\mathrm{alloc},m}}
+ \underbrace{\log \sum_{i \in m} \exp\!\left(q^{\!\top} \epsilon_i^{(m)}\right)}_{\Delta_{\mathrm{intra},m}}
- \log Z,
\label{eq:decomp}
\end{equation}
where $Z = \sum_{m'}\sum_{j\in m'} \exp(q^{\!\top} k_j^{(m')})$ is the shared attention normalizer. For audio-video comparison, $Z$ cancels:
\begin{equation}
\log \frac{\mathrm{mass}_a(q)}{\mathrm{mass}_v(q)}
= q^{\!\top}(c_a-c_v)
+ \left[
\log \sum_{i\in a}\exp(q^{\!\top}\epsilon_i^{(a)})
-
\log \sum_{j\in v}\exp(q^{\!\top}\epsilon_j^{(v)})
\right].
\label{eq:ratio_decomp_main}
\end{equation}
The first term is an architectural cross-modal allocation factor induced by shared-softmax attention; the bracketed term aggregates within-modality residual structure. A shared top-K rule therefore couples allocation with token-level selection before any model-specific empirical bias is considered.

\begin{figure*}[t]
    \centering
    \includegraphics[width=\textwidth]{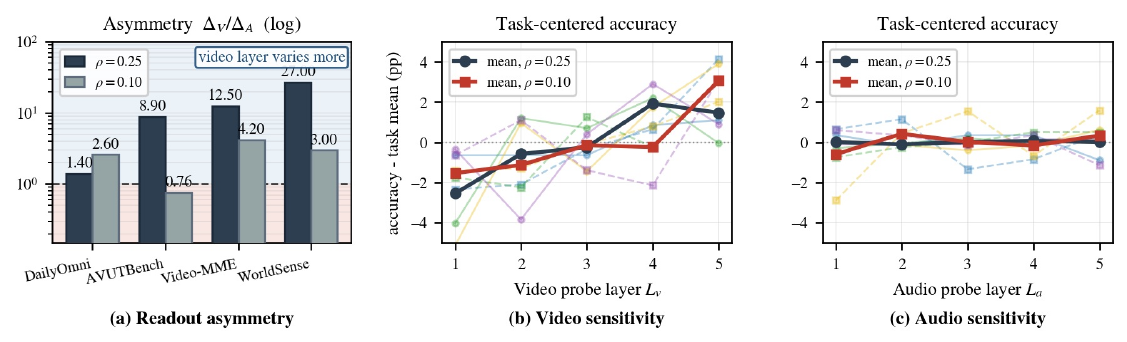}
       \vspace{-1.5em}
    \caption{\textbf{Audio and video become readable at different depths.}
    We sweep one modality's probe layer while fixing the other to
    \textsc{Macer}'s default.
    \textbf{(a)} Video readout is more layer-sensitive than audio readout in most settings.
    \textbf{(b)} Video readout performs best at L4-L5.
    \textbf{(c)} Audio readout remains relatively stable across layers.}
     \vspace{-1em}
    \label{fig:readout_asymmetry_main}
\end{figure*}

\paragraph{Observation 1: shared attention is not a neutral cross-modal saliency space.}
\label{sec:obs1}
\label{sec:audio_prior}
Even when the query explicitly demands video evidence, attention still allocates more mass to audio per token. At L3 under video-essential (VE) queries, every pooled query token in our diagnostic satisfies $\cos(q,c_a)>\cos(q,c_v)$ ($645/645$), placing the query direction in the audio half-space. Across all 28 layers, $\log(\mathrm{mass}_a/\mathrm{mass}_v)$ remains positive under both VE and audio-essential (AE) queries (Fig.~\ref{fig:teaser_audio_advantage}): query semantics shift the ratio in the expected direction, but do not overturn it.

Null-content controls argue that this is not merely a positional sink. Replacing the original audio with silence or scale-matched white noise nearly removes the audio advantage; mismatched real audio produces an intermediate shift (Appendix~\ref{app:null_content}). The shared attention is therefore responsive to audio content, but content-responsive audio mass is not the same as query-useful audio capacity. This is why raw shared attention is informative yet poorly calibrated for deciding a fixed audio-video retention budget.

Whether the saliency signal comes from audio, video, acoustic events, or shared attention over both modalities, the underlying assumption is often the same: one token-level signal can decide both \emph{how much} of each modality to keep and \emph{which tokens} to keep. These are different decisions. The audio-video balance is set by $q^{\!\top}(c_a-c_v)$ inside the decoder, encoder-side guides do not directly control this term, and shared ranking does not separate it from token-level scores. A global Top-$K$ rule turns token scores into modality budgets. Replacing the score changes which tokens rank highly, but still leaves the audio-video
split to be induced by ranking. The token budget should be allocated first, and local
selection applied afterward.

\paragraph{Observation 2: the audio-favoring bias is a budgeting problem, not a video-selection problem.}
In our matched-budget diagnostic, a content-aware within-video 
selector (ToMe-style merging) and a content-blind selector (random dropping) 
differ by at most 0.4 pp in audio mass share at the same video keep ratio, 
whereas reducing the video keep ratio itself increases audio's mass share by 
9.0 pp (Fig.~\ref{fig:allocation_vs_intra}b). The retained audio share also 
has direct accuracy consequences: a video-heavy split outperforms an 
audio-heavy split by 3.1 points (Fig.~\ref{fig:allocation_vs_intra}c). 
Cross-modal allocation must therefore be set at the capacity level, before 
within-modality selection begins.

\paragraph{Observation 3: audio and video become selection-readable at different depths.}
\label{sec:obs3}
Audio and video residuals become legible at different layers, and any single probe depth is misaligned with at least one modality (Fig.~\ref{fig:readout_asymmetry_main}). Video readout is layer-sensitive: earlier layers are allocation-dominated, while L4-L5 better expose within-video discrimination. Audio readout is comparatively flat across shallow layers, since the audio branch does not have to exit a ``how much audio at all'' regime before ranking residuals. The pattern is asymmetric-an early stable audio band and a later shallow video preference-rather than a sharp optimum at any single layer.

\paragraph{Design constraints implied by the mechanism.}
Together, these observations motivate \textsc{Macer}'s three core components. Observation 1 motivates allocation-normalized modality scoring, which removes modality-mass scale before local token ranking. Observation 2 motivates capacity-coordinated splitting, which assigns audio/video budgets before token selection. Observation 3 motivates modality-specific shallow readout, using an early stable audio probe and a later shallow video probe. Therefore, an Omnimodal compressor should separate capacity allocation, modality-local selection, and modality-specific readout.
\begin{figure*}[t]
    \centering
    \includegraphics[width=\textwidth]{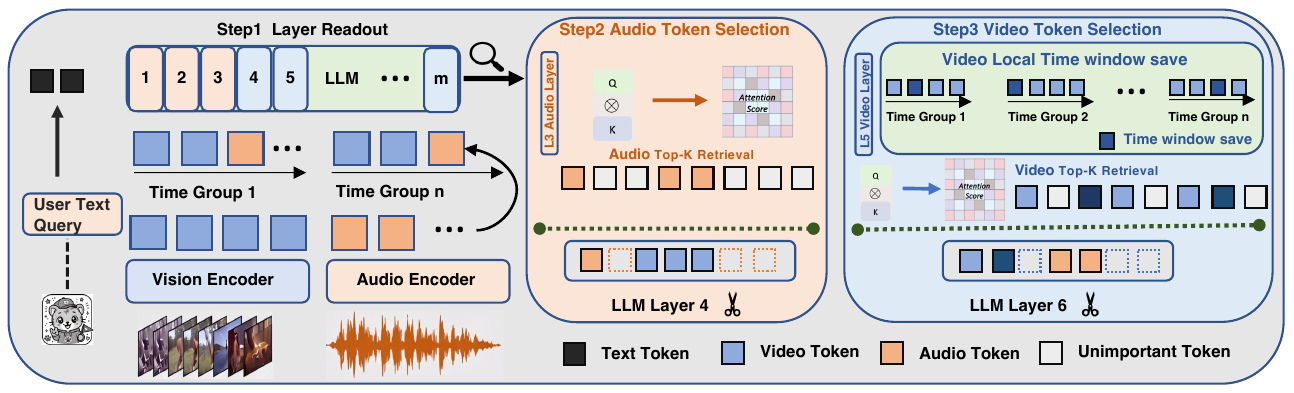}
    \vspace{-2em}
    \caption{\textbf{Macer pipeline.} \textsc{Macer} runs a single decoder prefill with staged
    modality-specific pruning. It reads audio attention at $\ell_a$ and immediately
    applies audio retention; the decoder then continues with retained audio and full
    video until $\ell_v$, where video retention is applied. The final retained audio
    and video tokens are merged with text tokens in the original sequence order, and
    the remaining decoder layers operate on the compressed sequence.}
    \vspace{-1em}
    \label{fig:macer_overview}
\end{figure*}

\section{Method: \textsc{Macer}}
\label{sec:method}
Section~\ref{sec:mechanism} shows that shared-decoder omnimodal compression is better treated as allocation before ranking. Shared attention
simultaneously allocates probability mass across modalities and ranks tokens
within each modality; a shared top-$K$ rule therefore lets the audio-favoring
allocation prior leak into token retention. \textsc{Macer}
(Modality-Aware Capacity-Explicit Retention) implements
the allocation-before-ranking principle by separating capacity allocation from within-modality selection. It first assigns audio and
video budgets explicitly, so the cross-modal capacity split is no longer an
accidental consequence of raw shared-attention mass. It then fills each assigned
budget with query-conditioned tokens selected only within the corresponding
modality. The actual pruning is staged within a single decoder prefill: audio is
pruned at $\ell_a$, and video is pruned later at $\ell_v$.

Concretely, \textsc{Macer} first assigns audio and video budgets explicitly, then fills each budget using modality-local, allocation-normalized scores read at modality-specific shallow layers. After the split, video retention uses a lightweight video-local coverage bonus as an auxiliary within-video correction for temporal concentration; this bonus never changes the audio--video capacity split. The pipeline is shown in
Fig.~\ref{fig:macer_overview}.

Let $I_a$ and $I_v$ denote the audio and video token sets entering the LLM, with
$n_a=|I_a|$ and $n_v=|I_v|$. Text tokens are always preserved; \textsc{Macer}
compresses only multimodal prefill tokens. Let $S_a$ and $S_v$ denote the retained audio and video token sets. Given keep ratio $\rho$, the
multimodal retention budget is

\begin{equation}
K_{\mathrm{mm}}=\min\{n_a+n_v,\lfloor \rho(n_a+n_v)+0.5\rfloor\}.
\end{equation}

Let $Q$ denote the question-token span after prompt templating, and let
$A_h^{(\ell)}(q,i)$ be the attention probability from query token $q$ to
multimodal token $i$ at layer $\ell$ and head $h$.

\textsc{Macer} is configured by modality-specific readout layers
$(\ell_a,\ell_v)$, a retained audio share $\alpha$, and a temporal coverage
weight $\lambda_c$. The audio budget fraction determines how the retained
multimodal budget is split between audio and video, while $\lambda_c$ weights
the coverage bonus used during video-token selection. The values used in the
experiments are reported in Sec.~\ref{sec:exp_setup} and
Appendix~\ref{app:calibration}.

\paragraph{Modality-specific shallow readout.}
Audio and video become useful for token selection at different shallow depths (Sec.~\ref{sec:mechanism}). Audio saliency is already stable in early layers, whereas video saliency benefits from a later shallow layer before pruning. \textsc{Macer} therefore reads audio attention at layer $\ell_a$ and video attention at layer $\ell_v$, rather than forcing both modalities to share one probe depth.

Pruning is applied in modality-specific stages at the corresponding readout layers. The audio and video budgets $(K_a,K_v)$ are computed once from the original multimodal token counts before either pruning stage. After layer $\ell_a$ is computed, \textsc{Macer} computes the allocation-normalized audio scores, selects the retained audio set $S_a$ under budget $K_a$, and removes the unretained audio tokens. The decoder then continues with all text tokens, the retained audio tokens, and the full video token set until layer $\ell_v$. After layer $\ell_v$ is computed, \textsc{Macer} scores video tokens, applies the video-local coverage rule under the video budget $K_v$, and removes the unretained video tokens.

\paragraph{Allocation-normalized modality scoring.}
Raw shared attention is not a neutral token-importance score: it contains both a modality-level allocation factor and a within-modality selection signal (Sec.~\ref{sec:mechanism}). Allocation-Normalized Modality Scoring (ANMS) removes the modality-level attention mass before ranking tokens within each modality. For modality $m\in\{a,v\}$, \textsc{Macer} normalizes each
query-to-token attention row within the target modality before averaging across query tokens and heads:
\begin{equation}
\widetilde{A}_{m,h}^{(\ell)}(q,i)
=
\frac{A_h^{(\ell)}(q,i)}{\sum_{j\in I_m}A_h^{(\ell)}(q,j)+\varepsilon},
\qquad
s_i^{(m)}=
\frac{1}{|Q|H}\sum_{q\in Q}\sum_{h=1}^{H}
\widetilde{A}_{m,h}^{(\ell_m)}(q,i).
\label{eq:macer_score}
\end{equation}
Audio scores are normalized only over $I_a$, and video scores only over $I_v$. The two ranked lists are never pooled into a shared audio-video top-$K$. ANMS decides which audio tokens compete with other audio tokens and which video tokens compete with other video tokens; it does not decide how much capacity each modality receives. The row-wise normalization removes the modality-mass scale from each $(q,h)$ contribution before averaging, so the final score reflects modality-local token preference rather than raw cross-modal attention mass.

\paragraph{Capacity-coordinated split.}
The first role of \textsc{Macer} is to correct the cross-modal capacity error created by the shared attention prior. If audio and video tokens are ranked together, audio can consume a disproportionate fraction of the retained budget before video tokens are ranked within their own modality. Capacity-Coordinated Split (CCS) prevents this by assigning how many audio and video tokens may survive before any within-modality selection is applied. This step reserves modality
capacity; it does not yet decide which tokens are useful.

Given retained audio share $\alpha$, we set
\begin{equation}
\bar{K}_a=\lfloor \alpha K_{\mathrm{mm}}+0.5\rfloor,
\qquad
\bar{K}_v=K_{\mathrm{mm}}-\bar{K}_a,
\label{eq:capacity_nominal}
\end{equation}
and then reassign unused capacity if one modality has fewer available tokens than
its nominal budget:
\begin{equation}
K_a=\min\{n_a,\bar{K}_a+\max(0,\bar{K}_v-n_v)\},
\qquad
K_v=K_{\mathrm{mm}}-K_a.
\label{eq:capacity_split}
\end{equation}
Here $\alpha$ is a retained-capacity share, not a raw token-count share. This distinction matters because audio often contributes many raw tokens, while the shared attention already favors audio. Fixing $\alpha$ makes the audio-video capacity split an explicit deployment choice, rather than an implicit consequence of raw attention magnitude or a one-way guide from another modality.

\paragraph{Modality-local token retention.}
Capacity correction is necessary but not sufficient: assigning more budget to video only improves the answer if the recovered video slots contain evidence that the decoder can use. \textsc{Macer} therefore fills each modality's budget using the allocation-normalized scores from Eq.~\eqref{eq:macer_score}. Given scores $(s^{(a)},s^{(v)})$ and budgets $(K_a,K_v)$, audio tokens are retained by modality-local top-$K$:
\begin{equation}
S_a=\operatorname{TopKIdx}_{K_a}\{s_i^{(a)}:i\in I_a\}.
\label{Top-k}
\end{equation}
Video retention follows the same modality-local principle, with a lightweight coverage bonus to avoid concentrating all retained video tokens in the same temporal region. Let $c(j)$ denote the fixed temporal partition index of video token $j$, obtained from the model-provided temporal position id. Let $r_g=|\{u\in S_v:c(u)=g\}|$ be the number of already retained video tokens in partition $g$. Starting from $S_v=\emptyset$, at each greedy step we select
\begin{equation}
j^\star
=
\arg\max_{j\in I_v\setminus S_v}
\left[
    s_j^{(v)}
    +
    \sqrt{\frac{\lambda_c}{1+r_{c(j)}}}\;
\right],
\label{eq:chunk_coverage_greedy}
\end{equation}
then add $j^\star$ to $S_v$ and update the corresponding partition count. The first term keeps selection query-conditioned, while the second provides a diminishing bonus for less-covered temporal regions. Setting $\lambda_c=0$ recovers pure video top-$K$.

This rule operates only within the assigned video budget $K_v$; it does not change the audio-video capacity split and does not use audio evidence to guide video selection. Unlike asymmetric readout, ANMS, and CCS, the coverage bonus is an auxiliary video-side correction for temporal concentration and is ablated separately in Sec.~\ref{sec:main_results}. After selection, the retained multimodal tokens $S_a\cup S_v$ are merged with all text tokens in their original sequence order. The remaining decoder layers then operate on this compressed
sequence.
\section{Experiment Results}
\label{sec:experiments}

\subsection{Evaluation Setups and Implementation Details}
\label{sec:exp_setup}

\paragraph{Benchmarks and baselines.}
We evaluate Qwen2.5-Omni-7B and 3B~\citep{xu2025qwen25omni} on AVUTBench~\citep{yang2025avut}, DailyOmni~\citep{zhou2025dailyomni}, WorldSense~\citep{hong2025worldsense}, and Video-MME~\citep{fu2024videomme}, covering audio-grounded, audio-video joint, visual-dominant, and video-centric regimes. We compare against representative baselines under a unified Qwen2.5-Omni decoder-side token-accounting protocol: Random retention as a content-blind lower bound; FastV~\citep{chen2024fastv}, an LMM-internal shared-attention ranker; and OmniZip~\citep{tao2025omnizip}, the closest executable encoder-side audio-guided omnimodal compressor under our setting. To test whether \textsc{Macer} transfers beyond the Qwen2.5-Omni family, we additionally 
evaluate it on OmniVinci-9B~\citep{ye2025omnivinci} against 
shared top-$K$ ranking on AVUTBench, DailyOmni, and WorldSense; 
full results are in Appendix~\ref{app:omnivinci}.

\paragraph{Implementation and deployment configurations.}
We implement \textsc{Macer} on Qwen2.5-Omni-7B/3B and OmniVinci-9B using NVIDIA RTX 6000 Ada 48GB GPUs. Text tokens are always preserved; compression is applied only to decoder-side multimodal prefill tokens. We report performance and inference cost at three retention rates, $\rho \in \{25\%, 35\%, 45\%\}$. Decoding is greedy and FlashAttention-2~\citep{dao2023flashattention} is enabled throughout. \textsc{Macer} exposes the audio-video allocation as an explicit deployment variable. We use $(\ell_a,\ell_v,\alpha,\lambda_c)=(3,5,0.30,0.20)$ for Qwen2.5-Omni-7B and $(4,7,0.32,0.30)$ for Qwen2.5-Omni-3B. These settings are selected once on a held-out split disjoint from all evaluation sets, then frozen across benchmarks, query types, and retention ratios. Appendix~\ref{app:alpha_sweep} shows that this choice lies in a stable allocation band: allowing a per-benchmark oracle allocation improves the normalized average by only $0.92$ percentage points.

\subsection{Main Results}
\label{sec:main_results}
All benchmarks are run through LMMs-Eval~\citep{zhang2024lmmseval} on the full
evaluation sets under a unified protocol. Table~\ref{tab:pareto_dominance}
reports the headline efficiency comparison; Table~\ref{tab:main} then provides
the full retention-controlled benchmark results.

\begin{table}[!t]
\centering
\caption{\textbf{Pareto comparison on Qwen2.5-Omni-7B.} 
At 25\% retention, \textsc{Macer} forms a lower-FLOPs operating point than OmniZip at 45\%.}
\label{tab:pareto_dominance}
\vspace{1pt}
\small
\setlength{\tabcolsep}{6pt}
\renewcommand{\arraystretch}{1.15}

\begin{tabular}{lcccccc}
\toprule
Method & Retain & FLOPs & AVUT & DailyOmni & WorldSense & Avg(3) \\
\midrule
OmniZip & 45\% & 39.2\% & 63.07 & 59.82 & 44.61 & 55.83 \\
\rowcolor{black!8}
\textbf{\textsc{Macer} (Ours)}
& 25\% & 34.8\% & \textbf{63.67} & \textbf{62.07} & \textbf{44.80} & \textbf{56.85} \\
\midrule
$\Delta$ 
& -20pp & -4.4pp & +0.60 & +2.25 & +0.19 & +1.02 \\
\bottomrule
\end{tabular}
\vspace{-18pt}
\end{table}

\paragraph{Pareto comparison.}
Table~\ref{tab:pareto_dominance} shows the headline efficiency result. At 25\% retention, \textsc{Macer} uses 4.4pp fewer FLOPs than OmniZip at 45\% while improving Avg(3) by 1.02 points, with consistent per-benchmark gains on AVUTBench (+0.60), DailyOmni (+2.25), and WorldSense (+0.19).

\vspace{-10pt}
\begin{table}[H]
\centering
\caption{\textbf{Comparison on Qwen2.5-Omni-7B/3B across audio-grounded, audio-video joint, visual-dominant, and video-centric benchmarks.} The \textbf{best} result among token-pruning methods within each retained-ratio block (25\%/35\%/45\%) is in bold, and the \underline{second-best} 
is underlined. FastV$^\dagger$ is evaluated on H100 due to 
OOM on Ada6000 under the long multimodal context; its average 
excludes Video-MME, denoted by ``-''. }
\vspace{10.0pt}
\label{tab:main}
\footnotesize
\setlength{\tabcolsep}{5pt}
\renewcommand{\arraystretch}{1.03}
\vspace{-2pt}
\begin{tabular}{l cc cccc c}
\toprule
\multirow{2}{*}{Method} 
& \multicolumn{2}{c}{Settings} 
& AVUT & DailyOmni & WorldSense & Video-MME 
& \multirow{2}{*}{ Norm. Avg.} \\
\cmidrule(lr){2-3}
& Retained Ratio & FLOPs Ratio & Avg. & Avg. & Avg. & Avg. &  \\
\midrule

\multicolumn{8}{c}{\textit{Qwen2.5-Omni-7B}} \\
\midrule
Full Tokens & 100\% & 100\% & 64.5 & 62.1 & 46.8 & 66.0 & 100\% \\
Random  & 55\% & 49.2\% & 61.0 & 59.2 & 43.6 & 65.4 & 95.5\% \\
FastV$^\dagger$   & 50\% & 54.1\% & 58.4 & 59.8 & 44.3 & - & 93.8\% \\

\specialrule{0.65pt}{0.4pt}{0.4pt}

OmniZip & 45\% & 39.2\% & \underline{63.07} & \underline{59.82} & \textbf{44.61} & \underline{66.3} & \underline{97.5\%} \\
\rowcolor{black!8}
\textbf{\textsc{Macer} (Ours)}
& 45\% & 50.1\% & \textbf{64.72} & \textbf{63.32} & \underline{44.00} & \textbf{66.9} & \textbf{99.4\%} \\

\specialrule{0.65pt}{0.4pt}{0.4pt}

OmniZip & 35\% & 29.6\% & \underline{61.05} & \underline{59.40} & \underline{43.98} & \underline{65.7} & \underline{96.0\%} \\
\rowcolor{black!8}
\textbf{\textsc{Macer} (Ours)}
& 35\% & 42.2\% & \textbf{64.63} & \textbf{62.32} & \textbf{45.05} & \textbf{66.2} & \textbf{99.3\%} \\

\specialrule{0.65pt}{0.4pt}{0.4pt}

OmniZip & 25\% & 20.6\% & \underline{59.10} & \underline{57.39} & \underline{38.18} & \textbf{66.5} & \underline{91.6\%} \\
\rowcolor{black!8}
\textbf{\textsc{Macer} (Ours)}
& 25\% & 34.8\% & \textbf{63.67} & \textbf{62.07} & \textbf{44.80} & \underline{66.2} & \textbf{98.7\%} \\

\specialrule{0.9pt}{1pt}{1pt}

\multicolumn{8}{c}{\textit{Qwen2.5-Omni-3B}} \\
\midrule
Full Tokens & 100\% & 100\% & 62.2 & 60.2 & 46.4 & 62.6 & 100\% \\
Random  & 55\% & 46.1\% & 58.7 & 57.4 & 42.8 & 61.1 & 94.9\% \\
FastV$^\dagger$   & 50\% & 49.2\% & 55.9 & 58.0 & 44.4 & - & 94.0\% \\

\specialrule{0.65pt}{0.4pt}{0.4pt}

OmniZip & 45\% & 36.1\% & \underline{61.30} & \textbf{59.73} & \underline{45.20} & \underline{62.80} & \underline{98.9\%} \\
\rowcolor{black!8}
\textbf{\textsc{Macer} (Ours)}
& 45\% & 48.5\% & \textbf{62.13} & \underline{59.15} & \textbf{46.45} & \textbf{63.21} & \textbf{99.8\%} \\

\specialrule{0.65pt}{0.4pt}{0.4pt}

OmniZip & 35\% & 26.8\% & \underline{60.10} & \underline{58.06} & \underline{44.30} & \underline{62.70} & \underline{97.2\%} \\
\rowcolor{black!8}
\textbf{\textsc{Macer} (Ours)}
& 35\% & 41.0\% & \textbf{60.52} & \textbf{58.15} & \textbf{45.55} & \textbf{63.37} & \textbf{98.3\%} \\

\specialrule{0.65pt}{0.4pt}{0.4pt}

OmniZip & 25\% & 18.3\% & \underline{58.90} & \underline{56.47} & \underline{43.40} & \underline{62.60} & \underline{95.5\%} \\
\rowcolor{black!8}
\textbf{\textsc{Macer} (Ours)}
& 25\% & 34.1\% & \textbf{59.74} & \textbf{58.40} & \textbf{44.24} & \textbf{63.01} & \textbf{97.3\%} \\

\bottomrule
\end{tabular}
\vspace{-15pt}
\end{table}

\paragraph{Representative baseline comparison.}
Table~\ref{tab:main} shows broad matched-retention gains over compared token-pruning baselines, especially under aggressive compression. On Qwen2.5-Omni-7B, \textsc{Macer} beats OmniZip in 10 of 12 matched benchmark-ratio cells; at 25\% retention, the gains reach +4.57 on AVUTBench, +4.68 on DailyOmni, and +6.62 on WorldSense, with Video-MME preserved at full-token level. On Qwen2.5-Omni-3B, \textsc{Macer} wins 11 of 12 matched benchmark-ratio cells, including all four benchmarks at 25\% retention. On
OmniVinci-9B, \textsc{Macer} leads shared top-$K$ ranking by up to
12.9 points (Appendix~\ref{app:omnivinci}). A matched-retention comparison with concurrent DASH~\citep{li2026dash} appears in Table~\ref{tab:dash_appendix}. This widening-gap pattern reflects the same failure mode across baselines:  Random has no importance signal, FastV applies global Top-$K$ to decoder attention, and OmniZip uses an encoder-side audio-guided score but still lets the retained composition be induced by pruning. \textsc{Macer} allocates the audio-video token budget before selection.

\paragraph{Efficiency analyses.}
\label{sec:efficiency}

\begin{table}[!t]
\centering
\caption{\textbf{End-to-end inference efficiency on WorldSense} on a single NVIDIA RTX 6000 Ada GPU. FastV runs out of memory under the long context.}
\label{tab:efficiency_worldsense}
\vspace{1pt}
\footnotesize
\setlength{\tabcolsep}{5pt}
\renewcommand{\arraystretch}{1.05}

\begin{tabular}{l c c c c c c c}
\toprule
Method & $\rho$ & FLOPs & Acc. $\uparrow$ & Total Lat. $\downarrow$ & Prefill $\downarrow$ & Decode $\downarrow$ & Mem. $\downarrow$ \\
\midrule
Full Tokens & - & 100\% & 46.80 & 10.5s & $\sim$3.70s & $\sim$6.8s & 41G \\
FastV & 50\% & 54.1\% & \multicolumn{5}{c}{\cellcolor{orange!8}\textsc{OOM}} \\
\midrule
OmniZip & 45\% & 39.2\% & 44.61 & 9.54s & \textbf{2.38s} & 7.16s & 29G \\
\rowcolor{black!8}
\textsc{Macer} & 25\% & \textbf{34.8\%} & \textbf{44.80} & \textbf{9.17s} & 2.53s & \textbf{6.64s} & \textbf{25G} \\
\bottomrule
\end{tabular}
\vspace{-10pt}
\end{table}

Table~\ref{tab:efficiency_worldsense} confirms the Pareto efficiency advantage end-to-end. \textsc{Macer} compresses 20pp more aggressively than OmniZip (retaining only 25\% vs.\ 45\% of multimodal tokens), yet simultaneously achieves higher accuracy (+0.19), lower total latency (-0.37s), lower decoding cost (-0.52s), and lower peak memory (-4G), while using 4.4pp fewer FLOPs. \textsc{Macer} therefore Pareto-dominates OmniZip on accuracy, latency, memory,
and compute, while retaining far fewer tokens.

\paragraph{Ablation study.}
\label{sec:ablation}
Table~\ref{tab:ablation} isolates the contribution of each design component. Shared top-$K$ ranking couples allocation with selection, collapsing into a 57.6/42.4 audio/video split and capping accuracy at 54.81. Replacing it with \textsc{Macer}'s explicit capacity-coordinated split, modality-local scoring, and asymmetric readout fixes the budget at 30.0/70.0 and lifts accuracy to 56.54 (\textbf{+1.73}), showing that the main gain comes from decoupling allocation from selection. Adding the video-local coverage correction keeps the same budget and further raises accuracy to 56.85 (\textbf{+0.31}; \textbf{+2.04 over shared top-$K$}), indicating that coverage is a secondary within-video correction.

\begin{table}[!t]
\centering
\caption{\textbf{Ablation study on Qwen2.5-Omni-7B at 25\% retention.}
Retained A/V is the benchmark-macro retained multimodal composition, excluding text.
Coverage operates only within the assigned video budget and does not change the
audio/video split.}
\label{tab:ablation}
\vspace{1pt}
\footnotesize
\setlength{\tabcolsep}{3.5pt}
\renewcommand{\arraystretch}{1.08}

\begin{tabular}{l c c c c c}
\toprule
Variant & Retained A/V & AVUT & DailyOmni & WorldSense & Avg. \\
\midrule
Shared top-$K$ ranking
& 57.6 / 42.4
& 61.27 {\scriptsize(60.0/40.0)}
& 59.57 {\scriptsize(56.7/43.3)}
& 43.60 {\scriptsize(56.0/44.0)}
& 54.81 \\

MACER core, no coverage
& 30.0 / 70.0
& 63.33
& 61.82
& 44.47
& 56.54 \\

Single readout layer
& 30.0 / 70.0
& 57.47
& 60.87
& 44.80
& 54.38 \\

\rowcolor{black!8}
\textbf{\textsc{Macer} full}
& \textbf{30.0 / 70.0}
& \textbf{63.67}
& \textbf{62.07}
& \textbf{44.80}
& \textbf{56.85} \\
\bottomrule
\vspace{-25pt}
\end{tabular}
\end{table}

\section{Limitations}
\label{sec:Limitations}
\textsc{Macer} is designed for audio-video OmniLLMs that process audio and video tokens with a shared decoder. Extending the allocation-selection decomposition to models with substantially different multimodal fusion designs
remains future work. The method uses shallow decoder-side attention readout before pruning, which enables query-conditioned modality-local scoring. At the evaluated operating points, its latency and memory profile remains favorable in
our end-to-end measurements (Table~\ref{tab:efficiency_worldsense}). Broader modality combinations and additional backbone families are natural directions for follow-up work.

\section{Conclusion}
\label{sec:conclusion}
Omnimodal token compression should not be reduced to a single shared-ranking problem. In audio-video OmniLLMs that place both modalities in the same LLM-side attention stream, shared attention couples cross-modal capacity allocation with within-modality token selection, while audio and video become selection-readable at different shallow depths. \textsc{Macer} instantiates an allocation-before-ranking principle by assigning audio-video capacity explicitly, ranking tokens locally within each modality, and reading saliency at modality-specific shallow layers, all without retraining. The results on Qwen2.5-Omni-7B/3B and OmniVinci-9B support this decomposition as a design interface for future omnimodal compressors, where allocation, selection, and readout can be improved separately rather than collapsed into one shared score.

\clearpage

{
\small
\bibliographystyle{plainnat}
\bibliography{main}
}

\clearpage
\appendix

\section*{Appendix}

This appendix provides the experimental protocol, implementation details,
mechanism diagnostics, calibration and robustness sweeps, efficiency
accounting, and the additional cross-backbone study that support the main-paper claims. The main
paper now contains the primary benchmark table, headline efficiency table, measured
latency/memory table, and component ablation table. The appendix therefore keeps
only the extended evidence that is not needed for the main experimental narrative:
Appendix~\ref{app:protocol} describes the evaluation and calibration setup,
Appendix~\ref{app:method_details} gives the exact \textsc{Macer} procedure,
Appendix~\ref{app:mechanism} reports mechanism controls,
Appendix~\ref{app:calibration_ablation} reports calibration and robustness sweeps,
Appendix~\ref{app:efficiency} reports cost accounting,
Appendix~\ref{app:omnivinci} reports OmniVinci-9B transfer, and
Appendix~\ref{app:additional_visualizations} retains additional diagnostic
figures. Table~\ref{tab:appendix_roadmap} summarizes this appendix evidence map.

\begin{table}[H]
\centering
\small
\setlength{\tabcolsep}{3pt}
\caption{Mapping of main-paper claims to appendix sections. The appendix follows the main paper's scale-specific \textsc{Macer} settings.}
\label{tab:appendix_roadmap}
\begin{tabular}{p{3.6cm}p{1.9cm}p{6.5cm}}
\toprule
Main-paper claim & Appendix & Extended content \\
\midrule
Experimental setup and reproducibility & A & Benchmark taxonomy, preprocessing, diagnostic setup, calibration split, token statistics \\
\textsc{Macer} implementation & B & Exact algorithm, allocation-normalized scoring, explicit capacity split, video coverage rule \\
Shared ranking couples allocation and selection & C & Ratio decomposition, token-count term, Q-centroid diagnostic, packing and distance controls \\
Audio allocation is query-robust and content-conditioned & C & Per-layer attention statistics, frame-count sanity check, visualized silence/noise/mismatched-audio controls \\
Single-guide saliency does not directly control allocation & C & ToMe/random invariance, allocation-vs-ranking stress test, and concise encoder-side guide analysis \\
Audio/video readout is asymmetric & D & Centered readout sweeps, full $5{\times}5$ heatmaps, sensitivity ratios \\
Explicit capacity split and video coverage & D & Retained-share $\alpha$ robustness, temporal concentration, and matched DASH comparison; main Table~\ref{tab:ablation} reports the component ablation \\
Efficiency reporting & E & FLOPs-ratio estimates used for the efficiency tables; main Table~\ref{tab:efficiency_worldsense} reports measured latency \\
Cross-backbone support & F & OmniVinci-9B transfer under native packing and reused 7B readout \\
\bottomrule
\end{tabular}
\end{table}

\begin{table}[H]
\centering
\small
\setlength{\tabcolsep}{3pt}
\caption{Main evaluation-suite taxonomy. Grouping by modality demand helps interpret benchmark-dependent compression behavior, but is not used by \textsc{Macer} during inference.}
\label{tab:benchmark_taxonomy}
\begin{tabular}{p{2.1cm}p{3.4cm}p{2.4cm}p{4.1cm}}
\toprule
Benchmark & Modality demand & Temporal regime & Role \\
\midrule
AVUTBench & Audio-grounded & Mixed & Audio allocation sensitivity \\
DailyOmni & AV-joint & Short clips & Cross-modal grounding / event sequence \\
WorldSense & Visual-dominant, audio-augmented & Mixed/longer & Mixed-modality robustness \\
Video-MME & Video-centric & Short/med/long & Video-side anchor \\
\bottomrule
\end{tabular}
\end{table}

\section{Experimental Protocol and Reproducibility}
\label{app:protocol}

\subsection{Benchmarks and evaluation protocol}
\label{app:taxonomy}

We group benchmarks by the modality structure of answer-relevant evidence rather
than by dataset provenance. The grouping is used only to organize analysis; all \textsc{Macer}
configurations are fixed across benchmarks.

AVUTBench contains audio-grounded questions where the audio stream carries
primary answer evidence and the video stream often supplies scene context.
DailyOmni emphasizes audio--video event alignment, temporal ordering, and
cross-modal grounding. WorldSense is mostly visually driven but includes
non-trivial audio-augmented cases. Video-MME, evaluated in the no-subtitle
setting, provides a video-centric anchor for testing whether an omnimodal
compressor remains competitive when the primary evidence is visual.

We evaluate Qwen2.5-Omni-7B and Qwen2.5-Omni-3B under fixed preprocessing and
decoding settings for each benchmark. Text tokens are always preserved.
Compression is applied only to multimodal prefill tokens. Reported average scores
are benchmark-normalized against the corresponding full-token model, so
\textsc{Full} is 100\% for each benchmark before averaging.

\subsection{Model preprocessing and token statistics}
\label{app:preprocessing}

We distinguish dataset-level frame caps from decoder-side token counts. Dataset
preprocessing follows the benchmark protocols used in the main paper: Video-MME
permits up to 768 raw frames and WorldSense up to 128 raw frames before
model-side sampling. \textsc{Macer} itself operates only on the audio and video
tokens that enter the LLM after the model preprocessing pipeline. Unless a
benchmark-specific protocol explicitly changes the model sampler, the default
Qwen2.5-Omni video preprocessing uses 12 sampled frames, yielding $n_v=864$
decoder-side video tokens. Audio token counts vary with clip duration; under the
default preprocessing the median audio token count is approximately
$n_a\approx1460$.

This token-count imbalance is distinct from the retained capacity split. A
setting such as $\alpha=0.30$ intentionally allocates fewer retained tokens to
audio than its raw token share, counterbalancing the audio-favoring allocation
prior observed in the shared attention space. All keep ratios and retained
budgets are computed from decoder-side multimodal token counts, not from raw
frame caps.

\subsection{Mechanism diagnostic setup}
\label{app:diagnostic_setup}

The mechanism analyses use Qwen2.5-Omni-7B on paired audio--video examples. For
paired video-essential (VE) and audio-essential (AE) diagnostics, each clip is
associated with a VE query and an AE query when available. Attention statistics
are computed from question-text tokens after prompt templating, excluding system
tokens, BOS tokens, and the answer prefix. Unless otherwise stated, attention is
averaged over heads within each layer before computing modality-level summaries.

For the main mechanism study, we use $N=500$ paired examples. The layer-cell
diagnostic in Appendix~\ref{app:bias} and the null-content controls in
Appendix~\ref{app:null_content} use a smaller paired subset of $N=30$ clips
because those controls require paired interventions over original, silence,
white-noise, and mismatched-audio variants. Readout-layer sweeps evaluate
downstream accuracy under the same compression budget while varying audio and
video probe layers.

\subsection{Fixed default selection}
\label{app:calibration}

\textsc{Macer} uses fixed backbone-level defaults rather than benchmark-specific
or query-specific hyperparameters. We use a held-out development split of
approximately 100 clips, disjoint from all reported evaluation sets and balanced
across the modality-demand categories in Table~\ref{tab:benchmark_taxonomy}, to
finalize the backbone-level operating point. Once selected, the configuration is
frozen across all reported benchmarks, query types, and keep ratios within that
model scale. No model weights are updated.

Table~\ref{tab:macer_defaults} reports the scale-specific defaults used in the
main paper. The readout layers follow the shallow-readout diagnostics: audio is
read from an early stable layer and video from a later shallow layer.
Qwen2.5-Omni-7B uses $(\ell_a,\ell_v)=(3,5)$, while Qwen2.5-Omni-3B uses
$(\ell_a,\ell_v)=(4,7)$. These settings are frozen within scale; no
benchmark-specific or query-specific tuning is used for the reported results.

The retained audio share $\alpha$ is treated as a fixed retained-capacity
setting, not a raw audio-token share and not a per-example allocation policy. We
use $\alpha=0.30$ for 7B and $\alpha=0.32$ for 3B. These values instantiate the
same design rule---make the audio--video capacity split explicit rather than
letting shared attention determine it implicitly---but we do not assume that one
numerical value transfers unchanged across model scales. For video coverage, we
use $C=8$ coarse temporal chunks and scale-specific coverage strengths
$\lambda_c=0.20$ for 7B and $\lambda_c=0.30$ for 3B. All quantities in
Table~\ref{tab:macer_defaults} are fixed before the reported benchmark runs.
Appendix~\ref{app:alpha_robustness} further audits the 7B retained-audio
share and shows that the frozen $\alpha=0.30$ setting lies in a stable
operating band rather than requiring benchmark-specific allocation tuning.
\begin{table}[H]
\centering
\small
\setlength{\tabcolsep}{8pt}
\caption{Scale-specific \textsc{Macer} defaults used after held-out calibration. The settings follow the main paper: they are frozen within each model scale, but not shared across 7B and 3B.}
\label{tab:macer_defaults}
\begin{tabular}{lcc}
\toprule
Quantity & Qwen2.5-Omni-7B & Qwen2.5-Omni-3B \\
\midrule
Audio readout / pruning after $\ell_a$ & 3 & 4 \\
Video readout / pruning after $\ell_v$ & 5 & 7 \\
FLOPs proxy depth $L_p=\max(\ell_a,\ell_v)$ & 5 & 7 \\
Retained audio share $\alpha$ & 0.30 & 0.32 \\
Video chunks $C$ & 8 & 8 \\
Coverage strength $\lambda_c$ & 0.20 & 0.30 \\
Text-token handling & \multicolumn{2}{c}{Always preserved} \\
Model weights & \multicolumn{2}{c}{Frozen} \\
Forward passes & \multicolumn{2}{c}{Single prefill pass; cached readout rows} \\
\bottomrule
\end{tabular}
\end{table}

\subsection{Metrics and statistical tests}
\label{app:metrics}

Main benchmark tables report raw benchmark accuracy and the normalized
cross-benchmark average. For each benchmark, normalized accuracy is computed
relative to the full-token model under the same model scale and evaluation
protocol:
\begin{equation}
\mathrm{NormAvg}(m)=\frac{100}{|\mathcal{B}|}\sum_{b\in\mathcal{B}}\frac{\mathrm{Acc}_{m,b}}{\mathrm{Acc}_{\mathrm{Full},b}}.
\label{eq:normavg}
\end{equation}

Table~\ref{tab:macer_algorithm} summarizes the resulting single-forward-pass
\textsc{Macer} procedure.

\begin{table}[H]
\centering
\small
\caption{\textsc{Macer} algorithm. The procedure separates cross-modal capacity coordination from within-modality token selection. The readout layers, retained audio share, and coverage strength are scale-specific values from Table~\ref{tab:macer_defaults}.}
\label{tab:macer_algorithm}
\begin{tabular}{p{0.04\linewidth}p{0.90\linewidth}}
\toprule
1 & Set the total multimodal budget $K_{\mathrm{mm}}=\min\{n_a+n_v,\lfloor \rho(n_a+n_v)+0.5\rfloor\}$ and allocate audio/video budgets $(K_a,K_v)$ using the calibrated retained audio share $\alpha$. \\
2 & Run the decoder through layer $\ell_a$ on the full multimodal prefill and cache query-span attention rows for audio. \\
3 & Compute allocation-normalized audio scores $s^{(a)}$ using Eq.~\eqref{eq:appendix_score}; select audio tokens by modality-local top-$K$, $S_a=\mathrm{top}_{K_a}\{s_i^{(a)}:i\in I_a\}$, and prune unretained audio tokens. \\
4 & Continue the same forward pass with all text tokens, retained audio tokens, and full video tokens until layer $\ell_v$; cache query-span attention rows for video. \\
5 & Compute allocation-normalized video scores $s^{(v)}$, partition video tokens into $C$ coarse chunks, and initialize $S_v^{(0)}=\emptyset$ with chunk counts $n_c^{(0)}=0$. \\
6 & Run the score-aware chunk-coverage greedy rule in Eq.~\eqref{eq:appendix_greedy} for $K_v$ steps to obtain $S_v$. Setting $\lambda_c=0$ recovers pure video top-$K$. \\
7 & Continue the remaining decoder blocks on text tokens plus $S_a\cup S_v$ in the original sequence order. \\
\bottomrule
\end{tabular}
\end{table}

Mechanism intervention tests use paired sign tests unless otherwise stated. For
null-content controls, the paired unit is the clip, and the tested quantity is
the direction of the shift in mid-layer $\log(v/a)$ relative to the
original-audio condition.

Main accuracy tables are deterministic evaluations under fixed preprocessing,
fixed model weights, fixed decoding, and fixed hyperparameters. We do not attach
error bars to all benchmark cells unless repeated full evaluations or paired
per-example prediction files are available. For stochastic baselines such as
random token dropping, the random seed is fixed when computing the reported table
cells. Code and calibrated configurations will be released upon publication.

\section{\textsc{Macer} Implementation Details}
\label{app:method_details}

\subsection{Allocation-normalized modality scoring}

Raw shared attention is not a neutral token-importance score because it contains
both a modality-level allocation factor and a within-modality selection signal.
Allocation-Normalized Modality Scoring (ANMS) removes the modality-level
attention mass before ranking tokens within each modality. For modality
$m\in\{a,v\}$, \textsc{Macer} normalizes each query-to-token attention row within
the target modality before averaging across query tokens and heads:
\begin{equation}
\widetilde{A}_{m,h}^{(\ell)}(q,i)
=
\frac{A_h^{(\ell)}(q,i)}{\sum_{j\in I_m}A_h^{(\ell)}(q,j)+\varepsilon},
\qquad
s_i^{(m)}=
\frac{1}{|Q|H}\sum_{q\in Q}\sum_{h=1}^{H}\widetilde{A}_{m,h}^{(\ell_m)}(q,i).
\label{eq:appendix_score}
\end{equation}
Audio scores are normalized only over $I_a$, and video scores only over $I_v$.
The two ranked lists are never pooled into a shared audio--video top-$K$. ANMS
decides which audio tokens compete with other audio tokens and which video tokens
compete with other video tokens; it does not decide how much capacity each
modality receives. We compute the normalization denominator in fp32 with a small
positive $\varepsilon$ for numerical stability.

\subsection{Capacity-coordinated split}

Given keep ratio $\rho$, the multimodal budget is
\begin{equation}
K_{\mathrm{mm}}=\min\{n_a+n_v,\lfloor\rho(n_a+n_v)+0.5\rfloor\}.
\label{eq:appendix_budget}
\end{equation}
Given retained audio share $\alpha$, we set
\begin{equation}
\bar{K}_a=\lfloor\alpha K_{\mathrm{mm}}+0.5\rfloor,\qquad
\bar{K}_v=K_{\mathrm{mm}}-\bar{K}_a,
\label{eq:appendix_nominal}
\end{equation}
and reassign unused capacity if one modality has fewer available tokens than its
nominal budget:
\begin{equation}
K_a=\min\{n_a,\bar{K}_a+\max(0,\bar{K}_v-n_v)\},\qquad K_v=K_{\mathrm{mm}}-K_a.
\label{eq:appendix_capacity}
\end{equation}
Here $\alpha$ is a retained-capacity share, not a raw token-count share. This
prevents the audio--video budget from being implicitly determined by raw
shared-attention magnitude or by a one-way guide from another modality.

\subsection{Video score-aware chunk-coverage greedy}

The video coverage rule is applied only after the video budget $K_v$ has been
assigned by the capacity split. It does not change the cross-modal capacity split
and does not use audio evidence. Its role is limited to reducing
over-concentration in video token selection while keeping the selection
score-aware.

We divide video tokens into $C$ coarse temporal chunks, using the model's native
temporal grouping when available and equal-size token bins otherwise. Let
$c(j)\in\{1,\ldots,C\}$ denote the chunk index of video token $j$. Given the
assigned video budget $K_v$, initialize $S_v^{(0)}=\emptyset$ and $n_c^{(0)}=0$
for all chunks. At greedy step $r=1,\ldots,K_v$, select
\begin{equation}
j_r=\arg\max_{j\in I_v\setminus S_v^{(r-1)}}\left[s_j^{(v)}+
\sqrt{\frac{\lambda_c}{1+n_{c(j)}^{(r-1)}}}\right],
\label{eq:appendix_greedy}
\end{equation}
then update
\begin{equation}
S_v^{(r)}=S_v^{(r-1)}\cup\{j_r\},\qquad
n_{c(j_r)}^{(r)}=n_{c(j_r)}^{(r-1)}+1,
\end{equation}
with all other chunk counts unchanged. The final retained video set is
$S_v=S_v^{(K_v)}$. The first term in Eq.~\eqref{eq:appendix_greedy} preserves
query-conditioned video saliency. The second term gives under-covered chunks a
diminishing bonus: an empty chunk receives bonus $\sqrt{\lambda_c}$, and after
$n$ retained tokens from the same chunk the bonus becomes
$\sqrt{\lambda_c/(1+n)}$. Large score gaps are still decided by $s_j^{(v)}$; the
coverage term mainly biases near-ties toward chunks with fewer retained tokens.

\subsection{Text-token preservation and single-forward-pass staged pruning}

Text tokens are always preserved. MACER applies pruning in two
modality-specific stages within a single forward prefill. After layer $\ell_a$,
audio scores $s^{(a)}$ are computed and only the selected audio tokens $S_a$ are
kept; all text tokens and all video tokens remain active. After layer $\ell_v$,
video scores $s^{(v)}$ are computed on the current sequence, the video budget
$K_v$ is applied with the video-local coverage rule, and only the selected video
tokens $S_v$ are kept. After the second pruning stage, the active sequence
contains all text tokens plus $S_a \cup S_v$ in the original relative order.

Consequently, the 7B configuration prunes audio after L3 and video after L5,
while the 3B configuration prunes audio after L4 and video after L7. This is a
single-forward-pass implementation with staged pruning, not a single-shot
pruning step after $L_p=\max(\ell_a,\ell_v)$. In practice, the score in Eq.~\eqref{eq:appendix_score} is averaged over the 
last $K=4$ tokens of the question span $Q$ rather than the full $Q$. This 
restriction is applied identically to both the audio readout at $\ell_a$ and 
the video readout at $\ell_v$.

\section{Mechanism Diagnostics}
\label{app:mechanism}

\subsection{Allocation--selection decomposition and token-count control}
\label{app:ratio_decomp}

For a query token $q$, decompose each modality-$m$ key as
\begin{equation}
k_i^{(m)}=c_m+\epsilon_i^{(m)},
\end{equation}
where $c_m$ is the per-clip modality centroid and $\epsilon_i^{(m)}$ is the
within-modality residual. The usual attention scaling factor is absorbed into
$q$. For audio--video comparison, the shared softmax normalizer cancels:
\begin{equation}
\log\frac{\mathrm{mass}_a(q)}{\mathrm{mass}_v(q)}=
q^{\top}(c_a-c_v)+\left[\log\sum_{i\in a}\exp(q^{\top}\epsilon_i^{(a)})-
\log\sum_{j\in v}\exp(q^{\top}\epsilon_j^{(v)})\right].
\label{eq:appendix_ratio_decomp}
\end{equation}
The first term is the cross-modal allocation term; the bracketed term is the
difference between within-modality residual aggregation terms. The bracketed term
is not a pure token-relevance term because it also contains token-count and
residual-concentration effects. The main point is that raw shared attention
necessarily mixes a cross-modal mass term with modality-local variation, while a
shared top-$K$ rule uses the mixed quantity for token retention.

The residual aggregation term contains a token-count component:
\begin{equation}
\log\sum_{i\in m}\exp(q^{\top}\epsilon_i^{(m)})=
\log n_m+
\log\left(\frac{1}{n_m}\sum_{i\in m}\exp(q^{\top}\epsilon_i^{(m)})\right).
\label{eq:appendix_count_decomp}
\end{equation}
This separates token count from residual concentration. Under the default
12-frame setting, video contributes $n_v=864$ tokens, while audio has median
$n_a\approx1460$. Therefore, the audio-favoring attention pattern is not a
consequence of video having more raw tokens. Conversely, \textsc{Macer}'s
retained capacity split deliberately under-samples audio relative to its raw
token share to preserve sufficient video evidence under aggressive pruning.

\subsection{Q-centroid mechanism}
\label{app:q_centroid}

\begin{figure}[H]
\centering
\includegraphics[width=\textwidth]{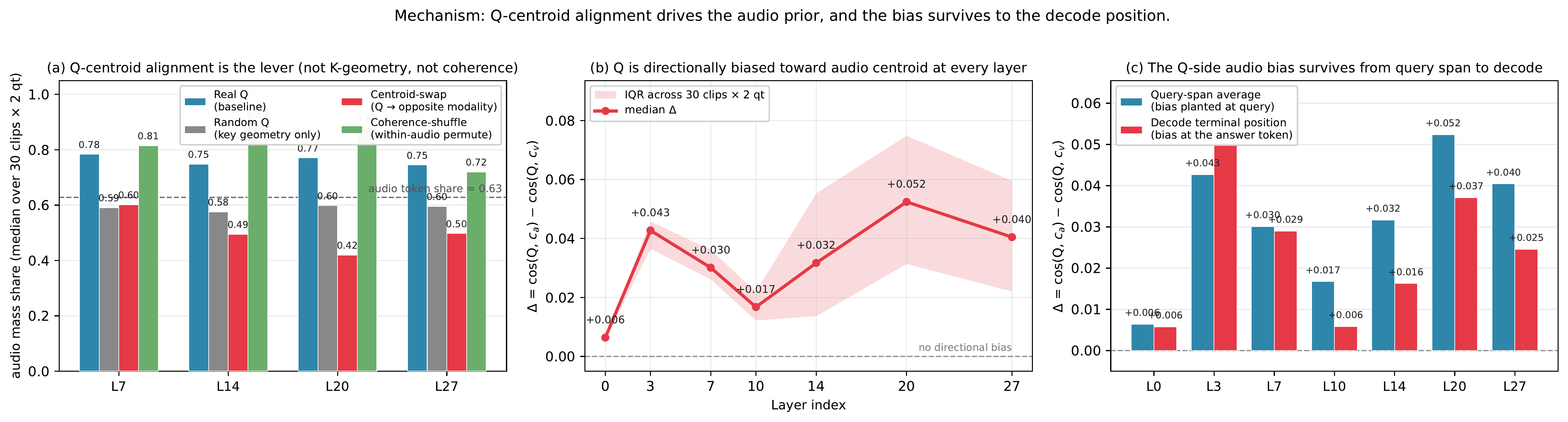}
\caption{\textbf{Q-centroid mechanism behind the audio prior.} (a) Query-side interventions show that audio mass is strongly controlled by alignment between the query direction and the modality centroids; key-side geometry and within-audio coherence controls alone do not explain the prior. (b) Across layers, the query representation is directionally closer to the audio centroid than to the video centroid. (c) The same Q-side audio bias is visible at the decode terminal position, indicating that the allocation tendency is not limited to the averaged query-token span.}
\label{fig:q_centroid}
\end{figure}

Equation~\eqref{eq:appendix_ratio_decomp} identifies $q^{\top}(c_a-c_v)$ as the
cross-modal allocation term. We therefore test whether the observed audio prior
is mainly driven by query-side alignment with the modality centroids, rather than
by key-side geometry alone or by within-audio coherence. Figure~\ref{fig:q_centroid}
reports three complementary diagnostics: interventions on the query direction,
layer-wise query-centroid alignment, and the persistence of the same bias at the
terminal decode position.

This diagnostic is consistent with the decomposition: the observed behavior is not
merely that audio keys are numerous or locally coherent, but that the shared
decoder places query directions in an audio-favoring centroid geometry. This is
why raw shared attention can remain a useful content-responsive signal while
still being a poor cross-modal capacity allocator.

\subsection{Packing structure and distance-matched controls}
\label{app:distance_matched}

To rule out a simple prefix-sink explanation, we inspect the input packing order.
The observed structure is
\begin{center}\small
\texttt{[system/user/template] $\to$ $\langle$vision\_bos$\rangle$ $\langle$audio\_bos$\rangle$ $\to$ V$_{288}$ $\to$ A$_{50}$ $\to$ V$_{288}$ $\to$ A$_{50}$ $\to$ V$_{288}$ $\to$ A$_{\mathrm{rest}}$ $\to$ [query]}
\end{center}
Audio is interleaved with video blocks rather than placed as a pure prefix. The
largest audio block appears close to the query, motivating the distance-matched
control below.

We control for token distance to the query using quantile-binned non-parametric
analysis. Table~\ref{tab:distance_matched} reports the equal-distance
audio--video ratio at selected layers. Early layers show that distance explains a
large part of the apparent audio advantage; mid-to-late layers retain an
audio-favoring residual even after distance matching.

\begin{table}[H]
\centering
\small
\caption{Distance-matched analysis. At early layers, matching token distance largely removes the audio advantage. Starting in the mid stack, a residual audio advantage persists after controlling for distance, indicating that the effect is not only a recency artifact.}
\label{tab:distance_matched}
\begin{tabular}{cccc}
\toprule
Layer & Equal-distance $\log(v/a)$ & $v/a$ ratio & Interpretation \\
\midrule
L7 & $-0.012$ & 0.988 & Positional bias dominates \\
L10 & $+0.007$ & 1.007 & Equal-distance parity \\
L14 & $-0.725$ & 0.484 & Audio residual emerges \\
L19 & $-0.886$ & 0.412 & Audio residual remains stable \\
\bottomrule
\end{tabular}
\end{table}

Distance matching and null-content ablation probe different confounds: distance
matching controls token position while keeping original content, whereas
mismatched-audio control disrupts semantic audio--video coupling while keeping a
real audio stream. At L14, the two estimates are numerically close:
distance-matched analysis gives $\log(v/a)=-0.725$, while the mismatched-audio
control gives $\log(v/a)=-0.746$. The close agreement between the distance-matched and mismatched-audio estimates
supports the same qualitative conclusion: the mid-layer residual is not explained
by either position or semantic alignment alone.

We also fit log-linear regressions
\[
\log(\text{attention})=\beta\log(\text{distance})+\epsilon
\]
per layer. The $R^2$ values range from 0.000 to 0.036, indicating that
attention--distance relationships are not well described by a simple power law.
The quantile-binned non-parametric analysis above is used for this reason and
does not rely on the regression model being well specified.

\subsection{Per-layer attention statistics}
\label{app:bias}

Table~\ref{tab:q2prime_layer} reports per-token attention statistics across all
28 decoder layers on the paired-query diagnostic subset (30 clips $\times$ 2
query types $\times$ 28 layers, head-aggregated, 1680 cells total).

\begin{table}[H]
\centering
\small
\caption{Per-layer audio--video attention statistics on selected layers. Video wins the per-token attention comparison in only 8.8\% of layer--clip--query cells pooled across query types and layers.}
\label{tab:q2prime_layer}
\begin{tabular}{cccc}
\toprule
Layer & $\log(v/a)$ median & $v>a$ rate & Interpretation \\
\midrule
L0 & $-3.02$ & 0/60 & Strong early positional bias \\
L3 & $-2.14$ & 0/60 & Strong early positional bias \\
L7 & $-0.73$ & 10/60 & Positional effect decaying \\
L9 & $-0.31$ & 13/60 & Weakest audio bias \\
L10 & $-0.42$ & 12/60 & Transition boundary \\
L14 & $-0.89$ & 7/60 & Audio residual emerging \\
L19 & $-1.03$ & 9/60 & Audio residual established \\
L27 & $-0.96$ & 9/60 & Final layer \\
\midrule
Pooled & $-0.96$ & 148/1680 (8.8\%) & \\
\bottomrule
\end{tabular}
\end{table}

Under video-essential queries, video wins in 12.1\% of cells (102/840). Under
audio-essential queries, video wins in 5.5\% of cells (46/840). The query shifts
the ratio systematically in the predicted direction, but does not reliably remove
the baseline preference for audio. This supports the main-paper phrasing
``query-robust'' rather than the stronger and less precise ``query-invariant.''

\subsection{Frame-count diagnostic}
\label{app:frame_sweep}

To check whether the 12-frame default creates the audio advantage, we vary the
video frame count on three paired clips. Table~\ref{tab:frame_sweep} shows that
the $v/a$ per-token ratio monotonically decreases with more video tokens in this
small diagnostic, suggesting that the 12-frame default is not unusually unfavorable
to video in this diagnostic.

\begin{table}[H]
\centering
\small
\caption{Frame-count diagnostic. The $v/a$ ratio decreases monotonically as frame count increases on the three tested paired clips, indicating that the 12-frame default is not obviously unfavorable to video among the tested settings. Higher frame counts were OOM on our hardware.}
\label{tab:frame_sweep}
\begin{tabular}{ccccc}
\toprule
Frames & $n_v$ & $v/a$ ratio & Monotonic pairs & Sign flips \\
\midrule
12 & 864 & 0.215 & -- & -- \\
16 & 1152 & 0.200 & 3/3 vs. 12 & 0 \\
20 & 1440 & 0.179 & 3/3 vs. 16 & 0 \\
\bottomrule
\end{tabular}
\end{table}

\subsection{Null-content controls}
\label{app:null_content}

\begin{figure}[H]
\centering
\includegraphics[width=\textwidth]{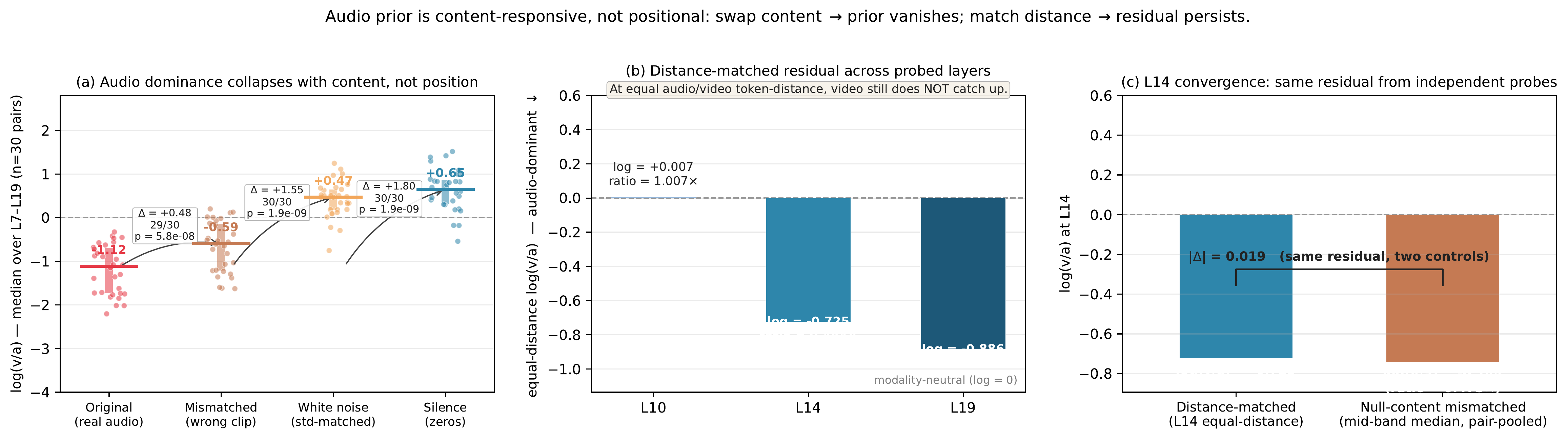}
\caption{\textbf{Null-content and distance-matched controls for the audio prior.} (a) Replacing original audio with mismatched real audio, scale-matched white noise, or silence shifts $\log(v/a)$ toward video; the prior therefore responds to audio content rather than behaving as an input-independent sink. (b) Equal-distance matching removes much of the early positional effect, but a mid-layer audio-favoring residual remains. (c) At L14, the mismatched-audio and distance-matched controls give nearly the same residual estimate, supporting the conclusion that neither recency nor semantic audio--video alignment alone explains the mid-layer effect.}
\label{fig:null_content_distance}
\end{figure}

We replace original audio with silence, scale-matched white noise, or mismatched
real audio. Silence and white noise nearly remove the audio advantage, while
mismatched audio produces an intermediate shift. This graded response shows that
audio attention is not an input-independent sink: the shared attention space is
responsive to audio content. Content responsiveness, however, does not make raw
attention magnitude a calibrated capacity signal under compression. Table~\ref{tab:null_content}
reports the corresponding shifts in mid-layer $\log(v/a)$ relative to the
original-audio condition, and Figure~\ref{fig:null_content_distance} visualizes
the same null-content and distance-matched controls.

\begin{table}[H]
\centering
\small
\caption{Shift in mid-layer $\log(v/a)$ relative to original audio across 30 paired clips. Silence and white noise produce large positive shifts, while mismatched audio produces a smaller but still positive shift.}
\label{tab:null_content}
\begin{tabular}{lcccc}
\toprule
Condition & Mean shift & Median shift & Min & Max \\
\midrule
Silence & $+1.83$ & $+1.80$ & $+1.17$ & $+2.81$ \\
White noise & $+1.60$ & $+1.55$ & $+0.81$ & $+2.49$ \\
Mismatched & $+0.52$ & $+0.48$ & $-0.04$ & $+1.66$ \\
\bottomrule
\end{tabular}
\end{table}

Sign tests against the null hypothesis of zero shift give silence (30/30 pairs,
$p=9.3\times10^{-10}$), white noise (30/30, $p=9.3\times10^{-10}$), and
mismatched audio (29/30, $p=2.9\times10^{-8}$). These tests support the direction
of the shift, and the tabled means and medians summarize the paired intervention
subset.

The original-to-mismatched shift (+0.52) reflects loss of semantic video--audio
alignment while preserving acoustic structure. The mismatched-to-white-noise
shift (+1.08) reflects destruction of semantic audio content. The
white-noise-to-silence shift (+0.23) reflects the residual contribution of
acoustic energy versus complete absence. The graded collapse supports a
content-conditioned interpretation rather than a pure structural-sink
explanation. One pair, clip c030, shows mismatched minus original = $-0.04$. We
retain it for transparency; the sign-test conclusion is unchanged.

We verify that null-content audio streams produce valid encoder outputs: token
counts are unchanged across conditions, output norms remain within a normal
range, and no NaNs or extreme values occur. The intervention therefore changes
audio content rather than breaking the input pipeline.

\subsection{Video-side compression and cross-modal balance}
\label{app:tome_random}

The main paper uses within-video merging and dropping as the empirical instance
of the broader claim that single-modality saliency does not explicitly control
cross-modal allocation. This diagnostic isolates video-side token selection from
retained video capacity.

\begin{figure}[H]
\centering
\includegraphics[width=0.88\linewidth]{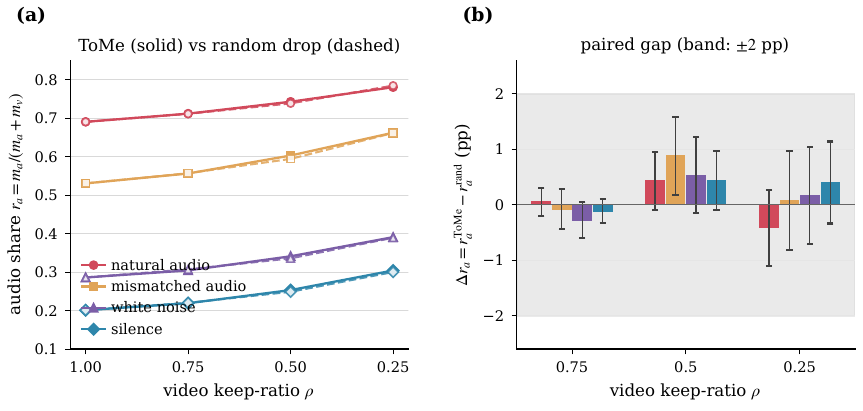}
\caption{\textbf{Within-video selector invariance at matched video budgets.} ToMe-style merging and random video dropping are compared at the same retained video ratios across natural-audio and null-content variants. (a) The audio share trajectories nearly overlap for ToMe and random dropping. (b) Paired differences remain within a narrow band, showing that the dominant cross-modal effect is retained video capacity rather than the local video selector used inside a fixed budget.}
\label{fig:tome_random_balance}
\end{figure}

For each keep ratio, we compare ToMe-style bipartite cosine merging on video
tokens against random video dropping. Both methods operate inside the video
modality only and use the same retained video budget. After compression, we
measure the resulting audio-to-video attention mass ratio in the downstream LLM
layers. Figure~\ref{fig:tome_random_balance} shows the matched-budget comparison
between ToMe-style merging and random video dropping.

This diagnostic isolates whether a content-aware within-video selector changes
cross-modal mass more than a content-blind selector when the retained video
budget is fixed. Similar trajectories indicate that the dominant cross-modal
effect is retained video capacity rather than local video-token identity. The
claim is not that within-video selection is irrelevant to accuracy, but that it
is not an explicit lever for audio--video allocation.

Compressing video can increase audio's relative mass share under the shared
normalizer. This effect does not mean the query has become more audio-relevant.
Rather, the retained evidence composition has changed while the shared softmax
still allocates probability mass over the remaining token set. This motivates
explicit capacity allocation: \textsc{Macer} decides how many audio and video
tokens are retained before applying within-modality ranking.

\subsection{Allocation removal and within-video ranking preservation}
\label{app:allocation_ranking}

\begin{figure*}[t]
\centering
\includegraphics[width=\textwidth]{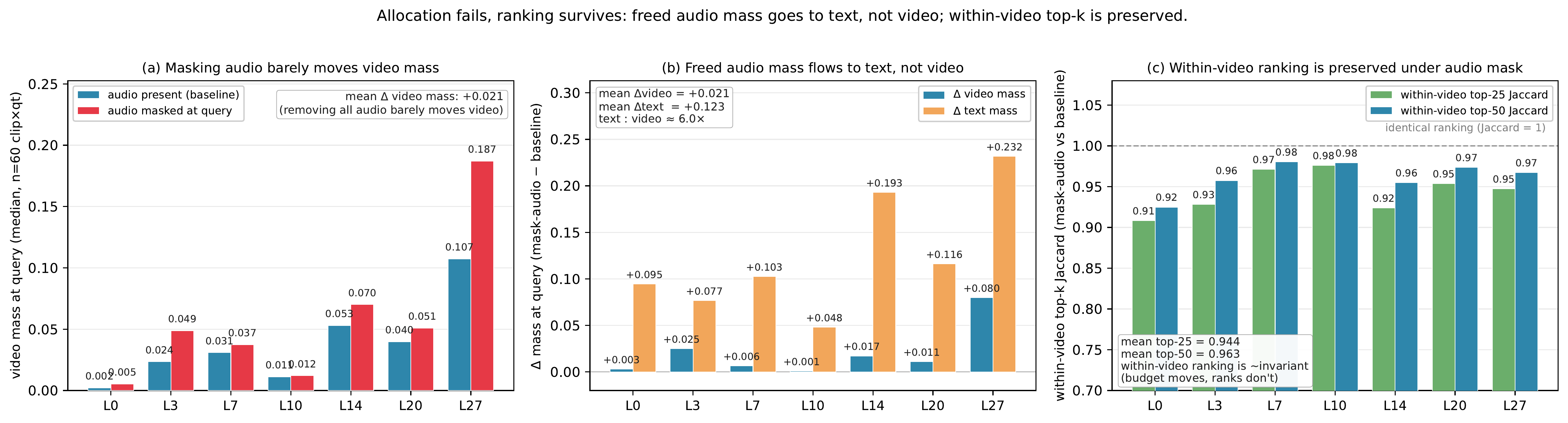}
\caption{\textbf{Allocation changes while within-video ranking largely survives.} (a) Masking audio at the query produces only a small increase in video attention mass. (b) Most freed audio mass flows to text rather than to video, indicating that removing the audio prior does not automatically restore video capacity. (c) Within-video top-K rankings remain highly stable under the audio mask. The diagnostic supports the separation used by \textsc{Macer}: cross-modal capacity allocation and within-modality ranking are distinct decisions.}
\label{fig:allocation_fails_ranking_survives}
\end{figure*}

We further separate cross-modal allocation from within-video ranking by masking
audio at the query while measuring both video mass and video-token ranking
stability. If the audio prior were simply preventing the model from finding video
tokens, removing audio should substantially increase video mass and change the
video top-$K$. Figure~\ref{fig:allocation_fails_ranking_survives} shows a
different pattern.

This stress test is diagnostic rather than a proposed compression method. It
shows that preserving or improving the ranking of video tokens is not enough when
the retained video budget itself is underspecified. A compressor must therefore
decide how many slots video receives before asking which video tokens should fill
those slots.

\subsection{Saliency replacement alone does not control global Top-$K$ allocation}
\label{app:one_way_guides}
Encoder-side guide methods use one modality, or a structure derived from one
modality, to decide retention in another modality or in shared temporal chunks.
This design can improve local token choice when the guide modality is well
aligned with answer evidence, but the guide remains a proxy for local salience
rather than a direct control over decoder-side audio--video capacity. By
Eq.~\eqref{eq:appendix_ratio_decomp}, audio--video attention mass contains an
allocation component $q^{\top}(c_a-c_v)$ in addition to within-modality residual
structure. Changing the guide direction changes which modality supplies the
proxy; it does not, by itself, create a controlled allocation variable.
\textsc{Macer} therefore delays compression until shallow query-conditioned
decoder evidence is available, assigns audio and video capacity explicitly, and
then performs token ranking only within each modality.

\section{Calibration and Ablations}
\label{app:calibration_ablation}

\subsection{Readout-layer sweep}
\label{app:readout_sweep}

This section reports the probe-layer analysis behind \textsc{Macer}'s asymmetric
readout choice. The centered sweep and heatmaps below are mechanism diagnostics
for the shallow readout behavior emphasized in the main paper. They support the
7B default $(\ell_a,\ell_v)=(3,5)$, while the 3B setting is calibrated separately
and reported in Table~\ref{tab:macer_defaults}.

For modality $m\in\{a,v\}$, define the layer-induced accuracy range
\begin{equation}
\Delta_m=\max_{\ell_m\in\{1,\ldots,5\}}\mathrm{Acc}(\ell_m)-
\min_{\ell_m\in\{1,\ldots,5\}}\mathrm{Acc}(\ell_m),
\end{equation}
where the other modality's probe layer is fixed to \textsc{Macer}'s 7B default.
The ratio $\Delta_v/\Delta_a$ measures whether video or audio readout is more
sensitive to probe-layer choice.

\begin{figure}[H]
\centering
\includegraphics[width=\textwidth]{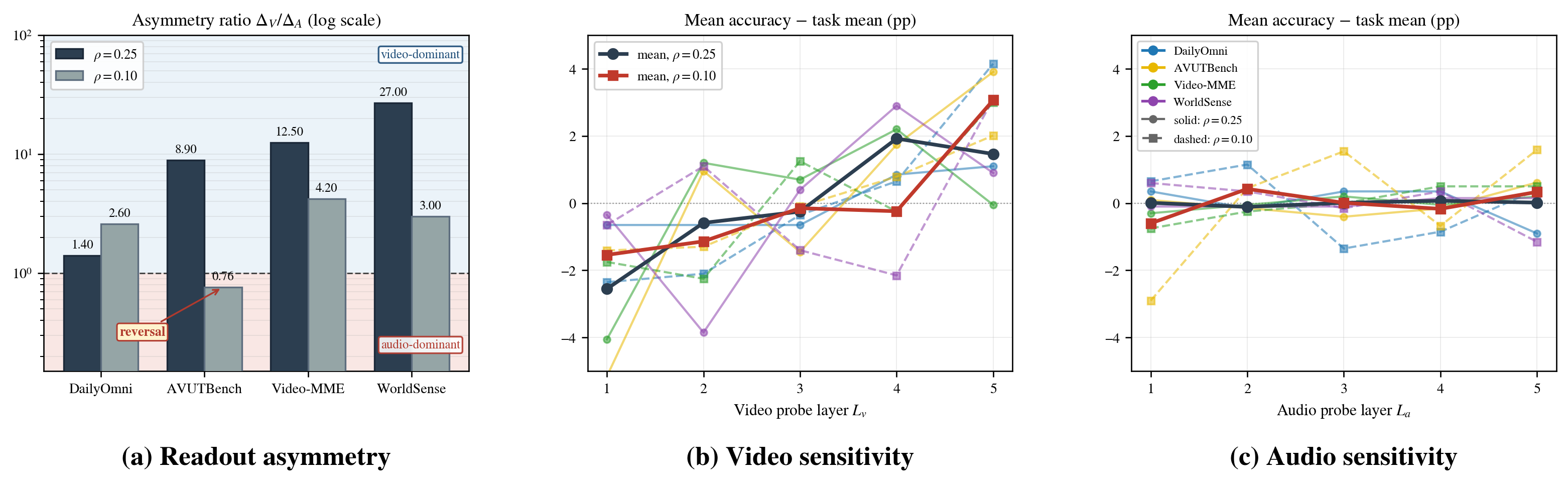}
\caption{\textbf{Readout-layer asymmetry between audio and video.} We sweep the probe layer of one modality while fixing the other modality to \textsc{Macer}'s 7B default readout layer, and report task-centered accuracy ($\mathrm{Acc}-\mathrm{mean}_{\ell}\mathrm{Acc}$) in percentage points. (a) The sensitivity ratio $\Delta_v/\Delta_a$ is above one in most benchmark--retention settings, showing that video readout is more layer-sensitive than audio readout. (b) Sweeping the video probe layer yields a consistent preference for later shallow layers, especially L4--L5. (c) Sweeping the audio probe layer produces small and inconsistent fluctuations around zero. These results motivate \textsc{Macer}'s asymmetric readout: an early stable audio probe and a later video probe.}
\label{fig:appendix_readout_asymmetry}
\end{figure}

The key conclusion is asymmetric sensitivity, not a sharp audio optimum. Video
readout changes downstream accuracy substantially as the probe layer moves,
whereas audio readout is comparatively flat across the shallow stack.
\textsc{Macer} therefore uses an early stable audio probe and a later video
probe, with exact layer choices calibrated per model scale. Figure~\ref{fig:appendix_readout_asymmetry}
summarizes the centered readout sweep used to diagnose audio--video readout
asymmetry.

\subsection{Full 5x5 readout heatmaps}
\label{app:readout_heatmaps}

\begin{figure}[H]
\centering
\includegraphics[width=\textwidth]{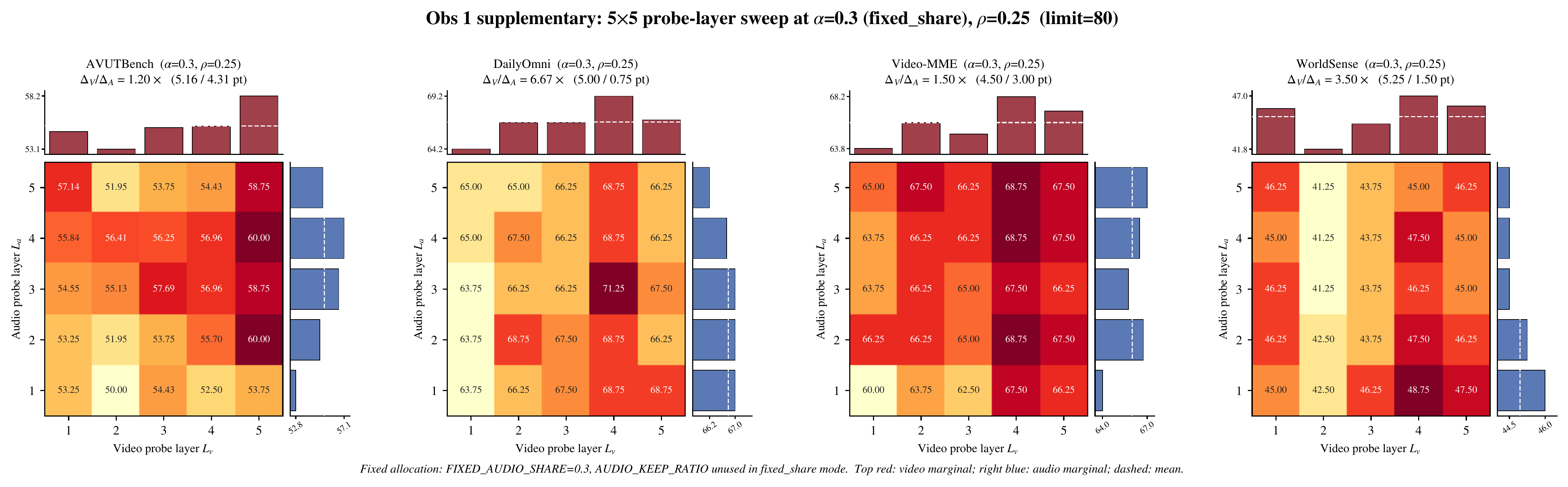}
\caption{Full $5\times5$ audio-video probe-layer sweep at the 7B calibrated retained-audio share $\alpha=0.30$ and representative keep ratio $\rho=0.25$. Rows vary the audio probe layer and columns vary the video probe layer under the same compression budget. Heatmaps use task-local color scales; numbers are raw scores. The fixed-share setting aligns the supplementary heatmap with the main-paper \textsc{Macer} configuration and supports the same asymmetry: video readout is more layer-sensitive than audio readout in most tasks.}
\label{fig:full_5x5}
\end{figure}

Figure~\ref{fig:full_5x5} expands the centered readout sweep to all $5\times5$
audio--video probe-layer pairs under the calibrated 7B capacity split.

\subsection{Asymmetry ratio across retention regimes}
\label{app:asymmetry_ratios}

Video sensitivity is strongest at the representative operating point $\rho=0.25$.
At the more aggressive $\rho=0.10$, the asymmetry weakens and can reverse on
AVUTBench. \textsc{Macer}'s default readout remains fixed across benchmarks and keep ratios
within each model scale; the reversal mainly explains why extremely tight
retention is less stable. The video-side sweep is consistent with the mechanism
analysis: video tokens become more useful for within-modality ranking in the
L4--L5 band, whereas earlier layers are more allocation-dominated. The audio-side
sweep does not show a sharp global optimum; it shows that shallow audio readout
is broadly stable. Table~\ref{tab:asymmetry_ratios} reports the corresponding
sensitivity ratios across retention regimes.

\begin{table}[H]
\centering
\small
\caption{Readout sensitivity ratio under two retention regimes. $\Delta_m$ is the accuracy range induced by sweeping modality $m$'s probe layer. Video readout is more sensitive than audio readout in most settings, especially at $\rho=0.25$. At $\rho=0.10$, the asymmetry weakens and reverses on AVUTBench, indicating that extremely tight budgets can change the relative importance of probe-layer choice.}
\label{tab:asymmetry_ratios}
\begin{tabular}{lcc}
\toprule
Benchmark & $\Delta_v/\Delta_a$ at $\rho=0.25$ & $\Delta_v/\Delta_a$ at $\rho=0.10$ \\
\midrule
AVUTBench & $8.91\times$ & $0.76\times$ \\
DailyOmni & $1.40\times$ & $2.40\times$ \\
Video-MME & $12.50\times$ & $4.29\times$ \\
WorldSense & $27.00\times$ & $9.00\times$ \\
\bottomrule
\end{tabular}
\end{table}

\subsection{Retained-audio share robustness}
\label{app:alpha_robustness}
\label{app:alpha_sweep}

Table~\ref{tab:alpha_robustness} reports a held-out robustness audit of the
retained-audio share at the representative aggressive setting $\rho=0.25$.
The sweep uses the full \textsc{Macer} procedure with the 7B readout and
coverage defaults, $(\ell_a,\ell_v)=(3,5)$ and $\lambda_c=0.20$, and varies only
$\alpha$. This audit is used to check whether the fixed retained-audio share is
brittle; it is not used to select benchmark-specific or query-specific
hyperparameters.

The deployed 7B setting $\alpha=0.30$ achieves the best normalized average among
fixed settings. The neighboring setting $\alpha=0.40$ is only 0.11 normalized
points lower, and a diagnostic oracle that selects $\alpha$ separately for each
benchmark is only 0.92 points higher. This supports interpreting $\alpha=0.30$
as a stable operating-band choice rather than a fragile benchmark-specific
setting.

At the default 12-frame setting, audio contributes more raw tokens than video
(median $n_a\approx1460$, $n_v=864$). Thus $\alpha=0.30$ is not a
token-count-matching choice. It deliberately retains fewer audio tokens than
audio's raw share, counterbalancing the audio-favoring allocation prior while
preserving enough audio capacity for audio-grounded and audio--video joint tasks.

\begin{table}[H]
\centering
\small
\setlength{\tabcolsep}{5.5pt}
\renewcommand{\arraystretch}{1.08}
\caption{\textbf{Robustness of the retained-audio share $\alpha$ on Qwen2.5-Omni-7B.} We sweep $\alpha$ at $\rho=25\%$ with full \textsc{Macer}, $(\ell_a,\ell_v)=(3,5)$, and $\lambda_c=0.20$ on a 200-example held-out robustness split spanning all four benchmark families. Norm. Avg. is computed by normalizing each row to the Full tokens row on the same split and then averaging across the four benchmark families. The deployed setting $\alpha=0.30$ achieves the best normalized average among fixed settings; the neighboring setting $\alpha=0.40$ is only 0.11 points lower, and a diagnostic benchmark-level oracle improves by only 0.92 points. Thus coarse task- or benchmark-conditioned allocation has limited headroom over the frozen deployment setting, supporting the interpretation of $\alpha$ as a stable operating-band choice rather than a brittle benchmark-specific hyperparameter.}
\label{tab:alpha_robustness}
\label{tab:alpha_sweep}
\begin{tabular}{lccccc}
\toprule
Setting & AVUT & DailyOmni & WorldSense & Video-MME & Norm. Avg. \\
\midrule
Full tokens & 65.83 & 69.50 & 44.00 & 77.00 & 100.00 \\
\midrule
$\alpha=0.10$ & 63.32 & 65.00 & \textbf{44.50} & 74.00 & 96.74 \\
$\alpha=0.20$ & 64.82 & 67.00 & 44.00 & 74.50 & 97.91 \\
$\alpha=0.30$ (frozen) & 66.50 & \textbf{68.50} & 43.50 & 74.50 & \textbf{98.80} \\
$\alpha=0.40$ & \textbf{67.00} & 68.00 & 43.00 & \textbf{75.00} & 98.69 \\
$\alpha=0.50$ & 64.00 & 66.50 & 42.50 & 74.00 & 96.40 \\
\midrule
Oracle per-benchmark $\alpha$ (diagnostic) & 67.00 & 68.50 & 44.50 & 75.00 & 99.72 \\
\bottomrule
\end{tabular}
\end{table}

\subsection{Coverage concentration and ablation}
\label{app:coverage_ablation}

We bin retained tokens into native 2-second temporal blocks and measure temporal
concentration with the Gini coefficient. Lower Gini indicates more uniform
timeline coverage. Table~\ref{tab:temporal_gini} reports the temporal
concentration diagnostic that motivates the video-local coverage term.

The no-coverage ablation sets $\lambda_c=0$, which recovers pure video top-$K$
under the same video budget $K_v$, the same modality-local video scores
$s^{(v)}$, and the same tie-breaking rule. Audio selection is unchanged and
remains modality-local top-$K$. Thus the ablation isolates the video-side
chunk-coverage term from the core allocation--selection separation.

\begin{table}[H]
\centering
\small
\caption{Temporal concentration of retained tokens under score-only selection. Video selection is more temporally concentrated than audio selection, motivating a video-specific score-aware chunk-coverage greedy rule.}
\label{tab:temporal_gini}
\begin{tabular}{lccc}
\toprule
Retention & Audio Gini & Video Gini & Interpretation \\
\midrule
$\rho=0.25$ & 0.06 & 0.33 & Video is substantially more concentrated \\
\bottomrule
\end{tabular}
\end{table}

Main Table~\ref{tab:ablation} reports the full component ablation. We keep the temporal-concentration diagnostic here to explain why the coverage term is included, but we do not duplicate the ablation table in the appendix.

\subsection{Matched-retention comparison with DASH}
\label{app:dash_comparison}

Table~\ref{tab:dash_appendix} provides a matched-retention comparison with the
concurrent DASH compressor. The comparison is included as an appendix baseline
check because it covers a subset of the full main-table grid. \textsc{Macer}
improves over DASH on five of six reported cells and is effectively tied on
WorldSense at 25\% retention.

\begin{table}[H]
\centering
\small
\setlength{\tabcolsep}{7pt}
\caption{Comparison with DASH~\citep{li2026dash} under matched retention. \textsc{Macer} improves over DASH on five of six cells and is effectively tied on WorldSense at 25\% retention.}
\label{tab:dash_appendix}
\begin{tabular}{lcccc}
\toprule
Benchmark & Retain & DASH & \textsc{Macer} & $\Delta$ \\
\midrule
AVUTBench  & 25\% & 60.90 & \textbf{63.67} & +2.77 \\
AVUTBench  & 35\% & 61.50 & \textbf{64.63} & +3.13 \\
WorldSense & 25\% & \textbf{44.90} & 44.80 & -0.10 \\
DailyOmni  & 25\% & 59.32 & \textbf{62.07} & +2.75 \\
DailyOmni  & 35\% & 60.65 & \textbf{62.32} & +1.67 \\
DailyOmni  & 45\% & 61.49 & \textbf{63.32} & +1.83 \\
\bottomrule
\end{tabular}
\end{table}

\section{Inference Cost Reporting}
\label{app:efficiency}

\textsc{Macer} prunes after shallow decoder-side readouts. For a conservative estimate, we charge full multimodal computation through the later modality readout layer, and charge the remaining decoder layers at the retained multimodal sequence length. This convention does not claim additional credit for the earlier audio-pruning stage.

\subsection{Token-layer cost proxy}
\label{app:token_layer_proxy}

Ignoring text tokens and constant factors, the normalized multimodal token-layer
cost is
\begin{equation}
\frac{(n_a+n_v)L_p+K_{\mathrm{mm}}(L-L_p)}{(n_a+n_v)L}
\approx
\frac{L_p+\rho(L-L_p)}{L}.
\label{eq:appendix_efficiency}
\end{equation}
Table~\ref{tab:token_layer_cost} reports the resulting proxy values for the
scale-specific pruning depths used in the main paper.

\begin{table}[H]
\centering
\small
\caption{Approximate multimodal token-layer cost under scale-specific pruning depths. These values are a cost proxy rather than wall-clock latency.}
\label{tab:token_layer_cost}
\begin{tabular}{ccc}
\toprule
Keep ratio $\rho$ & 7B: $L=28,L_p=5$ & 3B: $L=36,L_p=7$ \\
\midrule
0.10 & 26.1\% & 27.5\% \\
0.25 & 38.4\% & 39.6\% \\
0.35 & 46.6\% & 47.6\% \\
0.45 & 54.8\% & 55.7\% \\
\bottomrule
\end{tabular}
\end{table}

\subsection{FLOPs ratios used in the main tables}
\label{app:flops_accounting}

The main comparison table uses attention-corrected FLOPs accounting matched to
the baseline reporting protocol. Let $\eta$ denote the attention share of the
decoder computation. Input-prune methods use
\begin{equation}
\rho(1-\eta(1-\rho)),
\label{eq:appendix_input_flops}
\end{equation}
while probe-prune methods with a shallow full-context readout proxy use
\begin{equation}
\frac{L_p+(L-L_p)\rho(1-\eta(1-\rho))}{L}.
\label{eq:appendix_probe_flops}
\end{equation}
For \textsc{Macer}, we set $L_p=\max(\ell_a,\ell_v)$ when reporting estimated FLOPs. This is a conservative accounting proxy for the staged implementation because it charges all multimodal tokens until the later video readout rather than crediting the earlier audio-pruning stage. For Qwen2.5-Omni-7B we use $L=28$, $L_p=5$, and $\eta=0.235$. For Qwen2.5-Omni-3B we use $L=36$, $L_p=7$, and $\eta=0.36$, matching the scale-specific readout settings in Table~\ref{tab:macer_defaults}. Table~\ref{tab:flops_ratios_macer} gives the corresponding \textsc{Macer} FLOPs ratios.

\begin{table}[H]
\centering
\small
\caption{Attention-corrected \textsc{Macer} FLOPs ratios computed from Eq.~\eqref{eq:appendix_probe_flops}. The values are matched to the revised main-table FLOPs accounting.}
\label{tab:flops_ratios_macer}
\begin{tabular}{ccc}
\toprule
Keep ratio $\rho$ & 7B \textsc{Macer} FLOPs ratio & 3B \textsc{Macer} FLOPs ratio \\
\midrule
0.45 & 50.1\% & 48.5\% \\
0.35 & 42.2\% & 41.0\% \\
0.25 & 34.8\% & 34.1\% \\
\bottomrule
\end{tabular}
\end{table}

\subsection{Measured latency table moved to the main paper}
\label{app:latency}

Main Table~\ref{tab:efficiency_worldsense} reports the WorldSense latency and memory
measurement. We keep the FLOPs accounting in this appendix because it defines the
architecture-level cost estimate, but the measured wall-clock evidence now appears
in the main paper so reviewers can inspect it without consulting the appendix.

\section{Cross-Backbone Transfer Study}
\label{app:omnivinci}
\label{app:cross_backbone}

\paragraph{Additional backbone: OmniVinci-9B.}
To test whether \textsc{Macer}'s allocation--selection design is specific to
Qwen2.5-Omni, we further evaluate it on OmniVinci-9B~\citep{ye2025omnivinci}, an NVLabs omnimodal
backbone. We keep the shallow readout and one-shot pruning schedule identical to the
Qwen2.5-Omni-7B \textsc{Macer} configuration: audio scores are read at L3, video scores
at L5. Thus, the transfer study changes the backbone and
sweeps only the retained-audio share $\alpha$, while keeping the video coverage
strength fixed at $\lambda_c=0.20$. Table~\ref{tab:omnivinci_transfer} reports
the OmniVinci-9B transfer results.

\begin{table}[H]
\centering
\small
\setlength{\tabcolsep}{4.5pt}
\renewcommand{\arraystretch}{1.04}
\caption{\textbf{Cross-backbone transfer to OmniVinci-9B with explicit capacity split.}
DailyOmni and WorldSense use the corresponding full-token reference runs for normalization. 
Cons. Norm. Avg. normalizes each benchmark score to the corresponding full-token result and averages across AVUT, DailyOmni, and WorldSense. 
\textsc{Macer}-A ($\alpha=0.20$) yields consistent gains across all three retention levels, while \textsc{Macer}-B ($\alpha=0.40$) is strongest in the moderate-retention regime ($\rho\in\{35\%,45\%\}$), with mean $+7.50$ pp over shared top-$K$. 
Superscripts mark the full-token reference runs used for normalization.}
\label{tab:omnivinci_transfer}
\resizebox{\textwidth}{!}{%
\begin{tabular}{lcccccccc}
\toprule
Method & $\alpha$ & $\lambda_c$ & $\rho$ & AVUT & DailyOmni & WorldSense & Cons. Norm. Avg. & $\Delta$ vs. Shared \\
\midrule
Full tokens & -- & -- & 100\% & 64.36$^{\dagger}$ & 66.50$^{\ddagger}$ & 48.23$^{\ddagger}$ & 100.00 & -- \\
\midrule
Shared top-$K$ & -- & -- & 25\% & 55.78 & 53.75 & 41.34 & 84.40 & -- \\
\textsc{Macer}-A & 0.20 & 0.20 & 25\% & 55.78 & 48.75 & 45.93 & \textbf{85.07} & $+0.67$ \\
\textsc{Macer}-B & 0.40 & 0.20 & 25\% & 50.06 & 48.75 & 43.64 & 80.52 & $-3.88$ \\
\midrule
Shared top-$K$ & -- & -- & 35\% & 51.49 & 53.75 & 41.34 & 82.18 & -- \\
\textsc{Macer}-A & 0.20 & 0.20 & 35\% & 57.21 & 47.50 & 42.49 & 82.80 & $+0.62$ \\
\textsc{Macer}-B & 0.40 & 0.20 & 35\% & 55.78 & 48.75 & 44.79 & \textbf{84.28} & $+2.10$ \\
\midrule
Shared top-$K$ & -- & -- & 45\% & 52.92 & 52.50 & 40.19 & 81.50 & -- \\
\textsc{Macer}-A & 0.20 & 0.20 & 45\% & 60.07 & 46.25 & 43.64 & 84.45 & $+2.95$ \\
\textsc{Macer}-B & 0.40 & 0.20 & 45\% & 62.93 & 60.00 & 45.93 & \textbf{94.41} & $+12.91$ \\
\midrule
\multicolumn{9}{c}{\textit{Mean across $\rho \in \{25\%,35\%,45\%\}$}} \\
\midrule
Shared top-$K$ & -- & -- & mean & -- & -- & -- & 82.69 & -- \\
\textsc{Macer}-A & 0.20 & 0.20 & mean & -- & -- & -- & 84.11 & $+1.41$ \\
\textsc{Macer}-B & 0.40 & 0.20 & mean & -- & -- & -- & \textbf{86.40} & $+3.71$ \\
\midrule
\multicolumn{9}{c}{\textit{Mean across moderate retention, $\rho \in \{35\%,45\%\}$}} \\
\midrule
Shared top-$K$ & -- & -- & mean & -- & -- & -- & 81.84 & -- \\
\textsc{Macer}-A & 0.20 & 0.20 & mean & -- & -- & -- & 83.63 & $+1.79$ \\
\textsc{Macer}-B & 0.40 & 0.20 & mean & -- & -- & -- & \textbf{89.35} & $+7.50$ \\
\bottomrule
\end{tabular}}
\end{table}

The OmniVinci-9B study further strengthens our central claim. 
Under OmniVinci's native input packing, MACER again improves over shared-ranking compression on a second shared-decoder omnimodal backbone. 
This result provides evidence against two narrow explanations of our main findings: the gains are not merely a Qwen2.5-Omni artifact, and they do not rely on imposing a modified non-interleaved token layout. 
Instead, they show that explicitly exposing the audio/video capacity split can yield measurable gains even when the backbone uses a different omnimodal packing strategy. 
We keep this study in the appendix because it covers three benchmark families rather than the full four-benchmark main suite, but its role is important: it provides cross-backbone evidence that omnimodal compression benefits from separating capacity allocation from within-modality token selection.

\section{Additional Visualizations}
\label{app:additional_visualizations}

\subsection{PCA extended layers}
\label{app:pca_extended}

\begin{figure}[p]
\centering
\includegraphics[width=\linewidth]{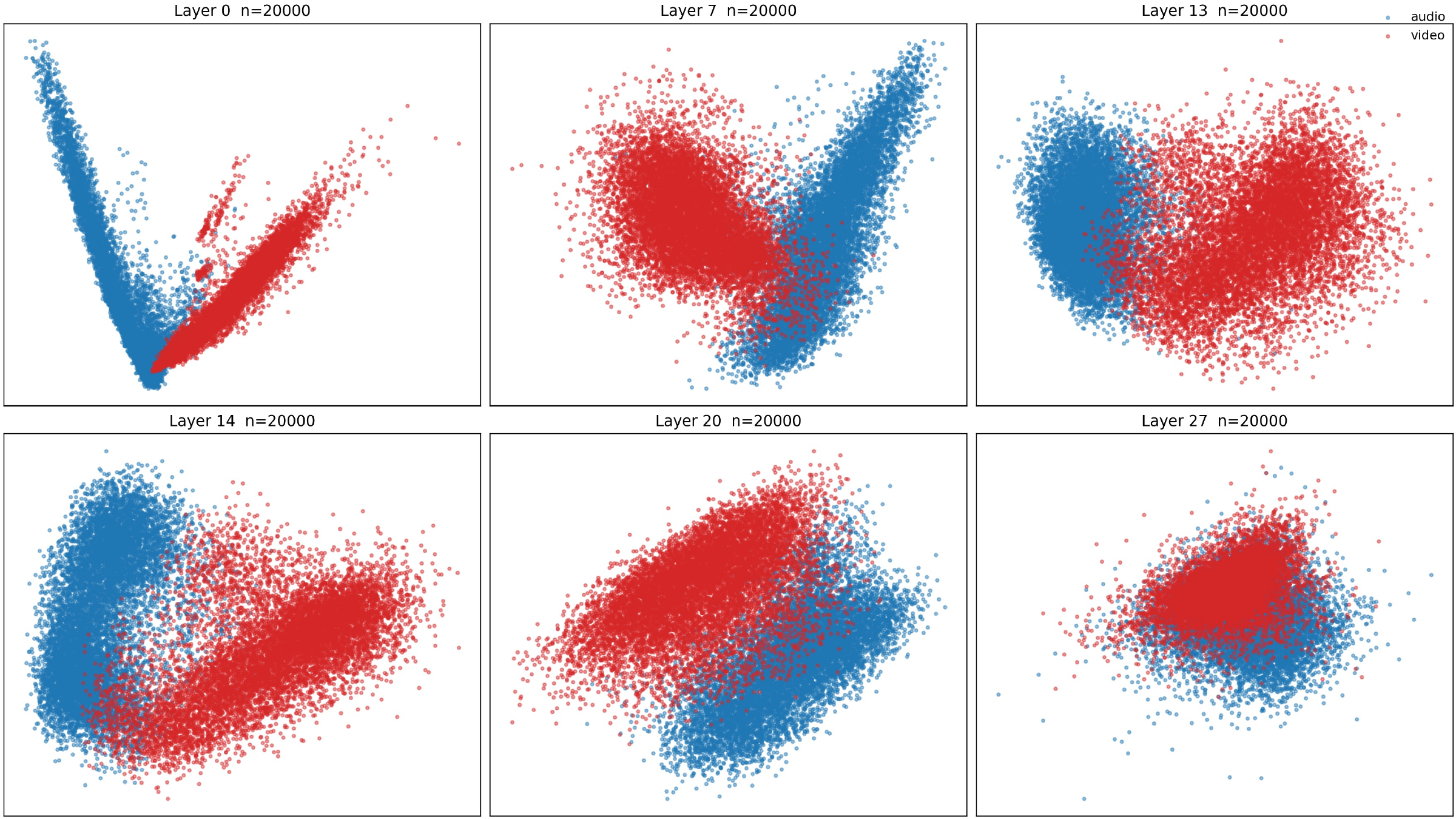}
\caption{PCA of query-key interaction space at layers $\{0,7,13,14,20,27\}$. Modality identity is a strong early-layer property and gradually mixes with depth.}
\label{fig:pca_extended}
\end{figure}

Figure~\ref{fig:pca_extended} shows PCA of the query-key interaction space at
additional layers beyond the main-paper teaser. The transition from sharp
modality separation in early layers to gradual mixing in later layers matches the
distance-matched and null-content evidence.

\end{document}